\documentclass[10pt,twocolumn,letterpaper]{article}

\usepackage[pagenumbers]{cvpr} 

\usepackage{graphicx}
\usepackage{amsmath}
\usepackage{amssymb}
\usepackage{booktabs}

\usepackage{subcaption}
\usepackage{amsmath,amssymb,amsfonts}
\usepackage{textcomp}
\usepackage{algpseudocode}
\usepackage{lmodern}
\usepackage{siunitx}
\usepackage{etoolbox}
\usepackage{array}
\newcolumntype{?}{!{\vrule width 1pt}}
\usepackage{mathtools}
\usepackage{multirow}
\usepackage{footmisc}
\usepackage{enumerate}
\usepackage{stmaryrd}
\usepackage{tabularx,ragged2e}
\usepackage[table]{xcolor}

\usepackage{mwe}
\usepackage{comment}
\usepackage{color}
\usepackage[linesnumbered,ruled,vlined]{algorithm2e}
\usepackage{xcolor}

\SetCommentSty{mycommfont}

\DeclareMathOperator*{\argmin}{arg\,min}
\usepackage{enumitem}
\setlist{nolistsep,leftmargin=*}

\usepackage[pagebackref,breaklinks,colorlinks]{hyperref}

\usepackage[capitalize]{cleveref}
\crefname{section}{Sec.}{Secs.}
\Crefname{section}{Section}{Sections}
\Crefname{table}{Table}{Tables}
\crefname{table}{Tab.}{Tabs.}

\def\cvprPaperID{5786} 
\def\confName{CVPR}
\def\confYear{2023}

\begin{document}

\title{Advancing Deep Metric Learning Through Multiple Batch Norms And Multi-Targeted Adversarial Examples}
\author{Inderjeet Singh\textsuperscript{1}\\
{\tt\small inderjeet78@nec.com}\\
\and
Kazuya Kakizaki\textsuperscript{1,2}\\
{\tt\small kazuya1210@nec.com}\\
\and 
Toshinori Araki\textsuperscript{1}\\
{\tt\small toshinori\_araki@nec.com}\\
\and
\textsuperscript{1}NEC Corporation\\
Kawasaki, Kanagawa, Japan\\
\and 
\textsuperscript{2}University of Tsukuba\\
Tsukuba, Ibaraki, Japan\\
}


\maketitle

\begin{abstract}
   Deep Metric Learning (DML) is a prominent field in machine learning with extensive practical applications that concentrate on learning visual similarities. It is known that inputs such as Adversarial Examples (AXs), which follow a distribution different from that of clean data, result in false predictions from DML systems. This paper proposes MDProp, a framework to simultaneously improve the performance of DML models on clean data and inputs following multiple distributions. MDProp utilizes multi-distribution data through an AX generation process while leveraging disentangled learning through multiple batch normalization layers during the training of a DML model. MDProp is the first to generate feature space multi-targeted AXs to perform targeted regularization on the training model's \textit{denser} embedding space regions, resulting in improved embedding space densities contributing to the improved generalization in the trained models. From a comprehensive experimental analysis, we show that MDProp results in up to $2.95\%$ increased clean data Recall@1 scores and up to $2.12$ times increased robustness against different input distributions compared to the conventional methods.
\end{abstract}

\section{Introduction}
\begin{figure*}[!h]
\centering
\includegraphics[width=0.95\linewidth]{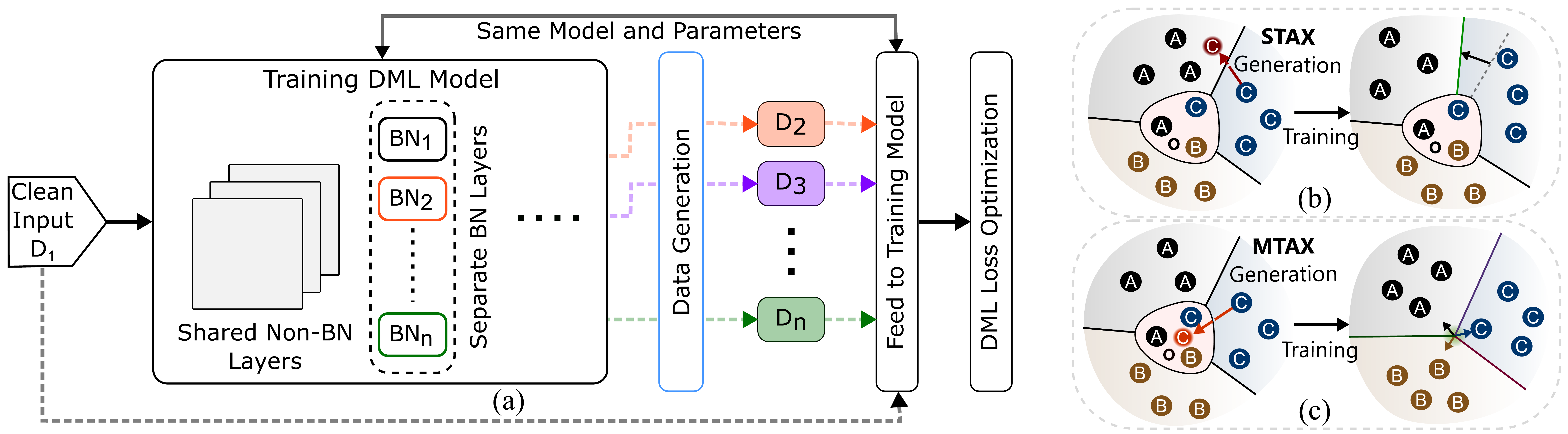}
\caption{(a) An outline of \textbf{MDProp} framework. MDProp generalizes the generation and the use of the inputs following multiple distributions while using separate BN layers. In particular, MDProp uses multi-targeted AXs (MTAXs) along with single-targeted AXs (STAXs), enhancing both accuracy and robustness. (b) The resulting feature space of the DML models trained using only STAXs. Because single-targeted AXs are not designed to target these overlapped regions, they leave the overlap unresolved. (c) MDProp generates MTAXs during training to specifically regularize the feature space overlap (region `O') in DML models, resulting in improved generalization in the trained DML models.}
\label{fig:main}
\end{figure*}

Deep metric learning (DML) has received considerable attention in recent years owing to its extensive applications, such as content-based information retrieval (CBIR), face recognition, voice recognition, and data dimensionality reduction. DML establishes similarities between objects by learning distance metrics in the feature space of deep neural networks (DNNs). Several techniques focusing on the DML model architecture, loss function, and data augmentation have been proposed to improve performance.

However, little attention has been paid to improving DML performance using input data with multiple types of adversarial examples (AXs) following different distributions. AXs are carefully crafted instances with small intentional perturbations following an adversary's defined generation process to fool a target DNN \cite{onepixel,pgd,fgsm,kurakin2018adversarial, sharif2019general, singh2022powerful}. Hence, reducing the performance and uncovering the restricted generalization ability of DML models.

To improve the generalization in DNNs for adversarial inputs, some studies focus on directly training DNNs using adversarial data. This technique is known as adversarial training \cite{kurakin2018adversarial,pgd,xie2019feature}. However, adversarial training improves only adversarial data performance at the expense of clean data performance degradation, which was later demonstrated as an inevitable accuracy-robustness tradeoff \cite{tradeoff}.

For \textit{end-to-end classification} DNNs, Xie et al. \cite{advprop} first challenged the accuracy-robustness tradeoff and discovered that clean data performance degradation in adversarial training occurs because the batch normalization (BN) layers in DNNs assume that input data follow a single distribution. They proposed using separate BN layers for the clean and adversarial data. AdvProp improved performance on the clean data, thus, enhancing generalization. But, the model performance for the adversarial inputs was not evaluated. Subsequent studies on AdvProp \cite{advprop} focus on self-supervised learning \cite{rst}, AdvProp's computational efficiency \cite{fastadvprop}, object detection \cite{chen2021robust}, and contrastive learning \cite{jiang2020robust,ho2020contrastive}.

We observe that there exists no method in DML that leverages multi-distribution data in the form of AXs while leveraging disentangled learning through multiple BN layers to improve the clean data performance and robustness against AXs of different kinds. Moreover, the AdvProp method proposed for end-to-end classification does not focus on \textit{multi-distribution data generation} or \textit{scaling} the use of separate BN layers.

This paper aims to improve the image retrieval performance of DML models on inputs with different distributions by considering clean and adversarial inputs. We propose a novel \textit{\textbf{M}ulti-\textbf{D}istribution \textbf{Prop}agation (MDProp) framework} (Fig. \ref{fig:main}(a)). MDProp first develops multi-targeted AXs (MTAXs) that follow different distributions than clean data and single-targeted AXs (STAXs). MDProp designs MTAXs to mimic the deep representations of multiple target classes, implying that they lie inside feature space regions with multi-class overlap. The use of MTAXs in training regularizes these overlapped feature space regions, as depicted in (b) and (c) of Fig. \ref{fig:main}. To handle the input distribution shift caused by MTAXs, MDProp scales the separate BN layer strategy followed by AdvProp \cite{advprop}. We also extend AdvProp to DML (hereafter called \textit{AdvProp-D}) by using an effective attack generation methodology specifically for DML.

\textit{To our knowledge, this work is the first instance in DML to simultaneously improve the performance on different input distributions using STAXs and MTAXs.} MDProp was evaluated thoroughly on standard DML benchmarks (\textit{CUB200-2011} (CUB200) \cite{cub200}, \textit{CARS196} \cite{cars}, and \textit{SOP} \cite{cars}), multiple ResNet \cite{resnet} architectures along with some recent state-of-the-art DML methods like S2SD \cite{roth2021simultaneous}, and different DML loss functions. We found up to 2.95\% improved R@1 scores for clean data while improving adversarial (out-of-distribution data) robustness by increasing R@1 scores up to 39.09\%. The improved performance on clean data and robustness results from better generalization capabilities and local Lipschitzness \cite{locallipschitzness} introduced due to the effective use of out-of-distribution features.

\section{Preliminaries}
\paragraph{DML.} DML aims to ﬁnd a distance metric $d_{\theta}:\psi \times \psi \mapsto \mathbb{R}$ on the feature space $\psi \subset \mathbb{R}^{D}$ of images $\mathcal{X}$ that best satisfy ranking losses \cite{multisimilarity,wu2017sampling,deng2019arcface,roth2020revisiting} deﬁned for class labels $\mathcal{Y}$. An adversary can conveniently create AXs to significantly affect a DML model's job \cite{sabour2015adversarial,lots,advrank,tolias2019targeted,zhou2021practical}.

\paragraph{AX Data Augmentation and Adversarial Training.} AXs added with adversarial noise $\delta$ bring additional features that help training DNN's parameters $\theta_{c}$ learn meaningful data representations \cite{tsipras2018there}. Adversarial training \cite{advtraining,kurakin2018adversarial,pgd,free_adv_training_shafahi2019,fast_adv_training_andriushchenko2020} is a straightforward strategy that incorporates AXs during training to make DNNs robust against AXs and noisy inputs \cite{yin2019fourier,zhang2019interpreting}, solving the following saddle point objective with a loss $\mathcal{L}$:
\begin{equation}
\min_{\theta_c}\mathbb{E}_{(x, y) \sim \mathbb{D}}\left[\max _{\delta \in \mathcal{S}} \mathcal{L}\left(\theta_c, x+\delta, y\right)\right],
\end{equation}
\noindent where $(x,y) \sim \mathbb{D}$ is the clean training data. $\delta$ is often crafted using first-order gradient-based methods \cite{pgd,fgsm}.

\paragraph{AdvProp.} To increase the accuracy of image recognition systems in an end-to-end \textit{classification} DL setting, AdvProp \cite{advprop} first proposed to use AXs during training. It was found that the \textit{input distribution shift} due to AXs was causing a reduced clean data accuracy for adversarial training because BN \cite{bn} assumes that all training samples $\mathcal{X}_{train}$ come from a single distribution $\mathbb{D}$. Hence, to leverage the adversarial features and handle the input distribution shift, AdvProp proposed using separate BN layers for the clean and adversarial data batches during training. The use of separate BN layers during training for input data with distinct distributions is called \textit{disentangled learning}.
For a DL classifier with parameters $\theta_c$, AdvProp optimized the following objective:
\begin{equation}
    \argmin_{\theta_c} \mathbb{E}_{\left(x,y\right) \sim \mathbb{D}} \left[\mathcal{L}\left(\theta_{c}, x, y\right) + \max_{\delta} \mathcal{L}\left( \theta_{c}, x+\delta, y\right) \right],
\end{equation}
\noindent where $(x,y)$ is a clean data instance with distribution $\mathbb{D}$, $\mathcal{L}$ is the \textit{classification} loss function, and $\delta$ is crafted for STAXs.

\paragraph{Multi-targeted AXs.}
When an AX can successfully be classified as multiple identities while being fed to a DL model during inference, it is called multi-targeted AX (MTAX). In the DML context, we generate MTAXs to fool a target model by imitating the deep features of target identities.  Let $f:\mathbb{R}^N \longrightarrow \mathbb{R}^D$ be a DML model with parameters $\theta$. For MTAX generation, an adversarial noise  $\delta_{f}^{m}$ is first crafted as:
\begin{gather}\label{eq: mtax noise generation}
    \delta_{f}^{m} = \argmin_{||\delta||_{\infty}\leq\epsilon} \left[ \frac{1}{T} \cdot \sum_{x^j\in S_B} ||f(x^{l}+\delta)-f(x^{j})||_p \right],
\end{gather}
\noindent where $x^l$ is the clean sample with identity $l$, $S_B$ is the batch sampled from training data $\mathcal{X}_{train}$ and contains images of target identities such that $j \neq l \ \ \forall x^j\in S_B$, $T$ is the number of impersonation targets, $\epsilon$ is the $l_{\infty}$-norm constraint on the size of adversarial noise $\delta$ to achieve the imperceptibility objective. Finally, the crafted $\delta_{f}^{m}$ is added to the clean sample $x^l$ and fools the target model $f$ even Eq. \ref{eq:mtax_fooling} partially holds.
\begin{equation}\label{eq:mtax_fooling}
    d(f(x^l+\delta_{f}^m),f(x^j)) \leq d(f(x^l),f(x_{r}^{l})) \ \ \ \forall x^j \in S_B,
\end{equation}
where $x_{r}^{l}$ is a gallery sample with same identity as $x^l$, and $d$ is a distance metric.

\section{Method}
\begin{algorithm*}[t]
\SetAlgoLined
\caption{Multi-Distribution Propagation Framework for $K$ input distributions}\label{alg}
\KwIn{Training data $(\mathcal{X}_{train}, \mathcal{Y}_{train})$, $f_{\theta_{n},\theta_{b}^{1},\theta_{b}^{2}...\theta_{b}^{K}}$, $\mathcal{B}$, $\phi_{n}$, $\phi_{b}$, $\mathcal{L}$, $\left\{\gamma^{k}, generator^k, \Lambda^{k}, T^{k}, \mathcal{L}^{k}\right\}_{k=2}^{K}$}
\KwOut{Trained Network $f$ with Parameters $\left\{\theta_{n}, \theta_{b}^{1}, \theta_{b}^{2}...\theta_{b}^{K}\right\}$}
\textbf{Initialize} parameters of $f$ as $\theta_{n},\theta_{b}^{1},\theta_{b}^{2}... \theta_{b}^{K}\leftarrow\phi_{n},\phi_{b},\phi_{b}+\gamma^{2},...\phi_{b}+\gamma^{K}$ \label{alg:line1}\\

\For{each training step \do}{
 Sample $(X,Y) \in (\mathcal{X}_{train},\mathcal{Y}_{train})$ of size $\mathcal{B}$ \label{alg:line3}\\
\For{k=2 to K \do}{
 \For{batch $(X,Y)$ \do}{
  Select targets $\mathbb{Y}^k=(Y_{1}^{'},Y_{2}^{'}...Y_{T}^{'})$ s.t. $y_{j}^{'}\neq y_{j}$ and $\{y_{j}^{'} \in Y_{i}^{'}, y_{j} \in Y\}$, $\forall i\in \{1,2...T\}$, $\forall j \in\{1,2...\mathcal{B}\}$ \\
  Sample $\mathbb{X}^{k} = \left\{ X_{i} \right\}_{i=1}^{T}$ s.t. all samples in $X_{i}$ has labels as $Y_{i}^{'}$ $\forall$ $i\in\{1,2...T\}$\\
  Generate $X^{k} = \left\{ x_i + \delta_i^m \right\}_{i=1}^{\mathcal{B}} = generator^{k}(f_{\theta_{n},\theta_{b}^{k}},\mathcal{L}^{k},X,Y,\mathbb{X}^k,\Lambda^{k})$ 
  \Comment{e.g., using Eq. \ref{eq: mtax noise generation} \& \ref{eq:mtax_fooling}}
 }
}
 Compute $loss = \sum_{\substack{x_i\in X \\ y_i \in Y}} \mathcal{L}(\theta_{n}, \theta_{b}^1, x_{i}, y_i) + \sum_{k=2}^{K}\sum_{\substack{x_i\in X^k \\ y_i \in Y}} \mathcal{L}(\theta_{n}, \theta_{b}^{k}, x_{i}, y_i)$  \\
 \Comment{While using separate BNs} \\

 \textit{Minimize} $loss$ by performing back-propagation to $\left\{\theta_{n}, \theta_{b}^{1}, \theta_{b}^{2}...\theta_{b}^{K}\right\}$
 }
\end{algorithm*}

We propose MDProp, a generalized framework for improving the image retrieval performance of DML models simultaneously on multiple input distributions. MDProp generates MTAXs, along with STAXs, as training data that follows different distributions. MDProp leverages the concepts of deep disentangled learning \cite{advprop} through multiple separate BN layers for each type of generated training data to better handle the input distribution shift.

Assuming $\theta = \{\theta_{n}, \theta_{b}\}$ are the trainable parameters of a conventional DML model $f$; where $\theta_{b}$ are BN parameters and $\theta_{n}$ are the remaining parameters. MDProp first generates multi-distribution data following different distributions than clean training data. The generated data is then combined with the clean data and used for training the model. During the whole data generation and training process, MDProp uses separate BN layers for each distribution of the generated data. The use of the separate BN layers is based on the concept of disentangled learning \cite{advprop}.  

For the original training data following $\mathcal{D}_1$ distribution, MDProp first generates predefined $K-1$ sets of data following $\mathcal{D}_2,\mathcal{D}_3,...\mathcal{D}_K$, while using a model $f_{\theta_{n},\theta_{b}^1,\theta_{b}^2,...\theta_{b}^K}$ with $K-1$ additional BN layers. The parameter set $\{\theta_{n},\theta_{b}^1\}$ is used for the clean data passes through the DML model $f$, and $\{\theta_{n},\theta_{b}^k\}$ is used for the generation and training pass of the data with  $\mathcal{D}_k$ distribution, for all $k\in \{2,3,..K\}$. Finally, MDProp optimizes the training objective $\mathcal{Z}_2$ as
\begin{equation}\label{eq:mdprop_objective}
    \mathcal{Z}_2 = \argmin_{\left\{\theta_{n},\theta_{b}^{1}...\theta_{b}^{K}\right\}}  \mathbb{E}_{\left\{\substack{(x,y)\sim \mathbb{D}^{k} \\ \forall k \in \{1,2...K\}}\right\}} \Big[ \mathcal{L}\left(\left\{\theta_{n},\theta_{b}^{k}\right\}, x, y \right)\Big],
\end{equation}

\noindent where $(x,y)$ is the training data, $\mathcal{L}$ is the DML training loss. MDProp is compatible with all popular DML loss functions. During inference, the auxiliary BN parameters $\{\theta_b^{2}, \theta_b^{3}...\theta_b^{K}\}$ are no longer required, and we only use $\theta_{test}^{m} = \{\theta_{n}, \theta_{b}^{1}\}$ parameters.

\subsection{Generation of Multi-Distribution Data}
MDProp uses a different data generation process to generate data with different distributions. Specifically, MDProp generates MTAXs along with STAXs using adversarial objectives and first-order optimization methods. MDProp crafts feature space adversarial noise $\delta_{f}^{m}$ as per Eq. \ref{eq: mtax noise generation} corresponding to a clean image $x$. The adversarial noise $\delta_{f}^{m}$ is then added to $x$, converting it into an MTAX. $\delta_{f}^{m}$ is distinctly generated for each clean image $x$ in the training batch.

Since the adversarial objectives of the MTAXs and STAXs are different, MDProp crafts and uses STAXs during training with added auxiliary BN layers. Later in the results, we will see that MDProp results in the best image retrieval performance when both MTAXs and STAXs, along with the clean data, are used for the training. MDProp is also compatible with conventional data augmentation methods \cite{cubuk2018autoaugment, lim2019fastautoaugment}.

\subsubsection{Why MTAXs?}
In the absence of a proper regularizer during the training of a DML model, there can be occurrences of feature space overlap (region `O' in Fig. \ref{fig:main}(b)) that reduces test time performance. Mathematically, the feature space of a subset $B$ of all classes $C$ will be considered to have an \textit{overlap} if there exists at least one input sample $x_{i}^{j}$ provided
\begin{equation}\label{eq:overlap}
    d_{\theta} \left( f\left( x_{i}^{j} \right), f \left( \Bar{x}^k \right) \right) \leq \tau \ \ \ \forall k\in B \ \ \& \ \ k\neq j ,
\end{equation}

\noindent where $\Bar{x}^k$ is the class center for the $k^{th}$ class, and $\tau$ is the classification threshold. When the deep features $f(x_{i}^{j})$ of an unperturbed input sample $x_i^j$ with identity $j$ lies inside the \textit{overlapped} region, it results in false ranking predictions from a DML model because Eq. \ref{eq:overlap} and Eq. \ref{eq:false_predictions} exist together.
\begin{equation}
\begin{aligned}
    d_{\theta} \left( f\left( x_{i}^{j} \right), f \left( \Bar{x}^k \right) \right) \leq \left( f\left( x_{i}^{j} \right), f \left( x_{g}^j \right) \right) \\ k\in B \ \ \& \ \ k\neq j.
\end{aligned}
\label{eq:false_predictions}
\end{equation}

\noindent Here $x_{g}^j$ is the \textit{gallery} or \textit{reference} image with identity $j$. The primary causes of the overlapped regions are high representation space similarity between the instances of different classes because of the low discriminative power of the trained DML model and limited training data complexity.

MDProp generates and uses MTAXs additional training data. The successful MTAXs generated during training lie in the overlapped regions. If the attack generation procedure is restricted due to computational budget, then the generated MTAXs during training may not end up inside overlapped regions but become closer to them. MDProp's effective use of these MTAXs induces a regularization effect. It pushes the model's parameters to transform the deep representation space, eliminating or lessening these overlapped regions (as illustrated in Fig. \ref{fig:main}), thereby improving generalization in the trained DML model.

\subsection{Working of MDProp}
The working of MDProp is shown in Algorithm \ref{alg} pseudo-code. MDProp first requires training data $\left(\mathcal{X}_{train},\mathcal{Y}_{train}\right)$, DML model $f_{\theta_{n},\theta_{b}^{1},\theta_{b}^{2}...\theta_{b}^{K}}$ with default hyperparameters, training batch size $\mathcal{B}$, pretrained $\{\phi_n,\phi_b\}$ parameters enabling transfer learning, noise set $\gamma$ for initializing $\theta_{b}^{k's}$, set $T$ which is the number of impersonation targets for the adversarial data generation, data generation recipe $generate^{k}$ with hyperparameters $\Lambda$, loss functions $\mathcal{L}^k$ with $k\in\{1,2,...K\}$ for training and data generation. The hyperparameters $\Lambda$ of $generate^{k}$ define generated attack's strength, the number of gradient update steps, and $L_\infty$ size constraints for the adversarial noise $\delta_{f}^{m}$ in the adversarial data generation process.

The next step is to initialize the model parameters $\left\{\theta_{n}, \theta_{b}^{1}, \theta_{b}^{2}...\theta_{b}^{K}\right\}$ using $\{\phi_n,\phi_b\}$. To enable transfer learning for the additional BN layers, noise $\gamma$ following an arbitrary distribution for compensating input distribution shift is optionally used. Thereafter, a batch of clean data $(X,Y)$ is sampled in each training step. Then $K-1$ data batches are generated using $generator^{k's}$ and separate BN layers, each following a different distribution. Finally, the $loss$ for all data batches is calculated. Each step ends with a back-propagation pass. The process is repeated to meet a predefined termination condition.

\subsection{AdvProp-D}
When MDProp uses a single additional BN layer, and the generated data is only STAXs, MDProp represents AdvProp-D, the DML extension of the AdvProp method \cite{advprop} proposed for the \textit{classification} DNNs. AdvProp can conveniently be applied to DML but generating \textit{effective} AXs during training is complex. Since the DML metrics are calculated at the embedding space, AdvProp-D generates AXs at the embedding space of the training model. Lastly, AdvProp-D optimizes the training objective as per Eq. \ref{eq:mdprop_objective}.

\section{Experimental Setting}
For a comprehensive assessment of \textit{MDProp}'s performance, we use various current baselines, DML architectures, benchmark datasets, and loss functions.

\noindent \textbf{Datasets.}
We use standard DML benchmarks: \textit{CUB200} \cite{cub200}, \textit{CARS196} \cite{cars}, and \textit{Stanford Online Product (SOP)} \cite{sop} datasets. We follow Roth et al. \cite{roth2021simultaneous} 
 to perform the pre-processing and train-test split. 

\noindent \textbf{Model architectures.}
To evaluate performance on models with varying capacities, we use \textit{ResNet50} \cite{resnet}, \textit{ResNet18} \cite{resnet}, and \textit{ResNet152} \cite{resnet} architectures. We use the publicly available \textit{ImageNet} \cite{imagenet} pre-trained parameters for transfer learning. We also use the state-of-the-art \textit{S2SD} method \cite{roth2021simultaneous} with \textit{ResNet50} architecture.

\noindent \textbf{Loss functions and model hyperparameters.}
We use the \textit{Multisimilarity} \cite{multisimilarity} and the \textit{ArcFace} \cite{deng2019arcface} losses for the training. We set DML model's embedding dimension $d=128$. For a fair comparison against baselines, we keep the remaining hyperparameters the same as Table 1's in \cite{roth2021simultaneous}.

\noindent \textbf{Attack hyperparameters during training.}
We use the well-known \textit{Projected Gradient Descent} (PGD) \cite{pgd} and \textit{Basic Iterative Method} (BIM) \cite{bim} for generating single and MTAXs. We set the number of iterations in PGD to 1, the PGD learning rate. To evaluate the effect of the size of adversarial noise, we set the $L^{\infty}$ constraint $\epsilon$ on the adversarial noise to $0.01$ and $0.1$. To understand the effect of MTAXs crafted for the different number of attack targets ($T$), we take $T=2,3,5,10$. The loss function for attack generation was kept squared $L^2$ norm.

\noindent \textbf{Baselines and ablation instances.}
We consider conventional training and adversarial training with targeted attacks as baselines. We compare MDProp against the baselines when using 2, 3, and 4 separate BN layers with STAXs and MTAXs for various \textit{numbers of attack targets}. We perform the comparisons for \textit{multiple datasets}, \textit{architectures}, and \textit{loss functions}. To validate the effect of separate BN layers, we evaluate training with MTAXs without auxiliary BN layers. 

\noindent \textbf{Evaluation metrics.}
We use the standard DML evaluation metrics: \textit{Recall@K} (R@K) \cite{recall} with $k=\{1,4\}$, \textit{Normalized Mutual Information} (NMI) \cite{nmi}, and $\pi_{ratio}$\footnote{See A in the appendix.}\cite{roth2021simultaneous}. Increased \textit{R@k} and \textit{NMI} values indicate improved image retrieval performance and clustering quality, respectively, and decreased $\pi_{ratio}$ values mean relatively increased inter-class and decreased intra-class distances in the embedding space.

\noindent \textbf{Robustness assessment of the trained models.}
To evaluate the effect of AXs on image retrieval performance, we generate \textit{single-targeted white-box AXs} corresponding to the clean samples in the test CUB200 \cite{cub200}, CARS196 \cite{cars}, and SOP \cite{sop} datasets. We use the \textit{PGD} \cite{pgd} update with 20 iterations, calling it \textit{PGD-20} attack. To change attack strength, we use 0.01 and 0.1 for the $\epsilon$ constraint. To save space, we only include the results for $\epsilon=0.01$, adding $\epsilon=0.1$ to supplementary material\footnote{See Table 6 in the appendix.}. Although, the result trends were found to be the same for both $\epsilon$ values. 

We keep the remaining attack hyperparameters the same as during the training. Because $\pi_{ratio}$ scores are averaged over all classes, making them prone to outliers and sometimes ignoring a few highly overlapped classes, they may not truly represent improved generalization constantly. Hence, we combine $\pi_{ratio}$ scores with the trained model's performance against MTAXs to conclude the improved generalization and eradication of the feature space overlap.

\section{Results and Discussions}

\setlength{\tabcolsep}{2pt}
\begin{table*}[t]
\begin{center}
\resizebox{\textwidth}{!}{
 & 
                                                                    \textbf{\textcolor{blue}{0.688}}  \\                 
                                                                    \bottomrule

\end{tabular}}

\caption{Image retrieval performance of models trained using \textit{Standard Training} (ST), \textit{Adversarial Training} (AT), \textit{AdvProp-D} (AP$^{'}$), and \textit{MDProp} (MP) with one ($\text{MP}^{'}$) and two ($\text{MP}^{''}$) additional BN layers, on the clean and STAX inputs from the CUB200 \cite{cub200} and CARS196 \cite{cars} datasets. T represents the number of attack targets used for MTAX generation during training. Adversarial datasets were generated using \textit{single-targeted white-box PGD-20} attacks with $\epsilon=0.01$ on the test sets \cite{roth2021simultaneous}. \textbf{\textcolor{blue}{Bluebold}} and \textbf{bold} denote best and second best results per setup.}
\label{table:1}
\end{center}
\end{table*}
\setlength{\tabcolsep}{1.4pt}

\noindent\textbf{Clean Data Performance.} Table \ref{table:1} presents results for the test \textit{CUB200} \cite{cub200} and \textit{CARS196} \cite{cars} datasets for the Multisimilarity \cite{multisimilarity} and ArcFace \cite{deng2019arcface} loss functions. \textit{MDProp} consistently and significantly outperformed not only the standard training and adversarial training baselines but also the AdvProp-D case. In particular, when MDProp used three BN layers with clean, STAX, and MTAX inputs, the image retrieval performance was the highest for the clean inputs. The performance of MDProp remained higher than that of AdvProp-D even when only MTAXs were used along with the clean data during training, indicating the significant impact of MTAXs. MDProp also exhibited similar performance gain patterns when used 4 separate BN layers\footnote{See D.4 in the appendix.}. Furthermore, we observed reduced $\pi_{ratio}$ scores for MDProp in most instances.

For larger SOP data also, MDProp trained using the mix of STAXs and MTAXs along with clean data performed the best for the clean inputs, which can be seen in Table \ref{table:sop_results}. However, the performance gains were relatively modest. We hypothesize that the reason for the lower gains on the SOP \cite{sop} data is the large number of classes with already low $\pi_{ratio}$ scores for the vanilla training baselines. A large number of classes increases the probability of ineffective adversarial target selection during MTAX generation. The already low $\pi_{ratio}$ scores for the vanilla training models indicate the presence of only a few overlapped embedding space regions for the data. 

\paragraph{Robustness Against Conventional AXs.} The benefits of MDProp are not limited to the improved image retrieval performance on unperturbed inputs following a single distribution. For the white-box adversarial inputs, Tables \ref{table:1}, \ref{table:r18_r152_r50s2sd_results}, and \ref{table:sop_results} show that MDProp results in significantly more robust DML models than the baselines by achieving up to 86\% higher recall scores, 47\% higher NMI scores, and a 41\% reduction in $\pi_{ratio}$ scores. When MDProp uses both STAXs and MTAXs with two additional BN layers during training, it results in the most robust models against STAXs, even compared with the AdvProp-D case. The robustness gains persisted even for the 4 separate BNs. This demonstrates the effect of the added generalization from the use of MTAXs with an additional BN layer.

\paragraph{Evaluating Feature Space Overlap.}
From Table \ref{table:1}, \ref{table:r18_r152_r50s2sd_results}, and \ref{table:sop_results}, we can see improved $\pi_{ratio}$ scores for MDProp in most of the instances\footnote{Check t-SNE plot in Fig. 4 of the appendix.}. We conclude the enhanced generalization by the use of MTAXs in MDProp by further evaluating the trained model's performance for the MTAX inputs and found that MDProp models trained using a mixture of STAXs and MTAXs perform the best\footnote{See Section D.1 in the appendix.}, followed by MDProp trained using MTAXs and STAXs, respectively.

\paragraph{Performance Across Architectures.} MDProp results in superior clean data performance and adversarial robustness when used with DL architectures of varying depth and even with the distillation-based S2SD \cite{roth2021simultaneous} method, as shown in Table \ref{table:r18_r152_r50s2sd_results}. Particularly, MDProp in the S2SD setting for the CUB200 dataset outperforms the state-of-the-art distillation-based standard training method \cite{roth2021simultaneous} by 1.39\% on R@1 score, and also on NMI and $\pi_{ratio}$ scores.

\paragraph{Effect of the Number of Attack Targets.}
From the experiments for $T$: 1, 2, 3, 5, and 10, we found that increasing $T$ improves performance on the clean data only up to a certain number for which the predefined \textit{generation} recipe's hyperparameters provide the sufficient semantic capability to the attack generation procedure to cause the positions in embedding space of generated MTAXs shift to the overlapped regions of the DML model under training\footnote{\label{footnote:all_tables}See D in the appendix.}. In particular, MDProp using clean and MTAXs performed the best for $T=3$, and MDProp using clean, STAXs, and MTAXs performed the best for $T=5$.
For smaller values of $T$, lesser performance improvements result because of the decreased probability of finding highly overlapped embedding space regions.

\setlength{\tabcolsep}{2pt}
\begin{table*}[t]
\begin{center}
\resizebox{\textwidth}{!}{
\begin{tabular}{lcrrrrrrrrrrrrrrrrrrr}
    \toprule
    \multirow{3}{*}{Method}& \multirow{3}{*}{$T$}& \multicolumn{9}{c}{ResNet50\cite{resnet}+S2SD\cite{roth2021simultaneous} Method} & & \multicolumn{4}{c}{ResNet18\cite{resnet}} && \multicolumn{4}{c}{ResNet152\cite{resnet}}\\\cmidrule{3-11} \cmidrule{13-16} \cmidrule{18-21}
    & & \multicolumn{4}{c}{Clean CUB200 Data} & & \multicolumn{4}{c}{\textit{Adversarial} CUB200 Data} & & \multicolumn{4}{c}{Clean CUB200 Data} & & \multicolumn{4}{c}{Clean CUB200 Data}\\\cmidrule{3-6} \cmidrule{8-11} \cmidrule{13-16} \cmidrule{18-21}
     &  & R@1 & R@4 & NMI & $\pi_{ratio}$ & & R@1 & R@4 & NMI & $\pi_{ratio}$ & & R@1 & R@4 & NMI & $\pi_{ratio}$ & & R@1 & R@4 & NMI & $\pi_{ratio}$\\ \toprule
    
    \rowcolor{gray!10}ST                                     & - &  \begin{tabular}{@{}c@{}} 67.69  \\ \footnotesize $\left[\pm 0.13 \right]$\end{tabular} & 
                                                                    \begin{tabular}{@{}c@{}} 86.32 \\ \footnotesize $\left[\pm 0.08 \right]$\end{tabular} &  
                                                                    \begin{tabular}{@{}c@{}} 71.46 \\ \footnotesize $\left[\pm 0.13 \right]$\end{tabular} & 
                                                                    1.123 & & 
                                                                    \begin{tabular}{@{}c@{}} 47.35 \\ \footnotesize $\left[\pm 1.24 \right]$\end{tabular} & 
                                                                    \begin{tabular}{@{}c@{}} 76.08 \\ \footnotesize $\left[\pm 0.64 \right]$\end{tabular} & 
                                                                    \begin{tabular}{@{}c@{}} 60.26 \\ \footnotesize $\left[\pm 0.40 \right]$\end{tabular} & 
                                                                    1.393 & & 
                                                                    \begin{tabular}{@{}c@{}} 58.81  \\ \footnotesize $\left[\pm 0.52 \right]$\end{tabular} & 
                                                                    \begin{tabular}{@{}c@{}} 81.34 \\ \footnotesize $\left[\pm 0.33 \right]$\end{tabular} &  
                                                                    \begin{tabular}{@{}c@{}} 66.12 \\ \footnotesize $\left[\pm 0.45 \right]$\end{tabular} & 
                                                                    1.131 & & 
                                                                    \begin{tabular}{@{}c@{}} 65.11 \\ \footnotesize $\left[\pm 0.28 \right]$\end{tabular} & 
                                                                    \begin{tabular}{@{}c@{}} 84.64 \\ \footnotesize $\left[\pm 0.10 \right]$\end{tabular} & 
                                                                    \begin{tabular}{@{}c@{}} 69.70 \\ \footnotesize $\left[\pm 0.02 \right]$\end{tabular} & 
                                                                    0.967  \\ \midrule
    \rowcolor{gray!10}AT                                     & 1 &  \begin{tabular}{@{}c@{}} 66.46 \\ \footnotesize $\left[\pm 0.59 \right]$\end{tabular} & 
                                                                    \begin{tabular}{@{}c@{}} 85.63 \\ \footnotesize $\left[\pm 0.12 \right]$\end{tabular} &  
                                                                    \begin{tabular}{@{}c@{}} 70.78 \\ \footnotesize $\left[\pm 0.40 \right]$\end{tabular} & 
                                                                    1.092 & & 
                                                                    \begin{tabular}{@{}c@{}} 45.13 \\ \footnotesize $\left[\pm 1.09 \right]$\end{tabular} & 
                                                                    \begin{tabular}{@{}c@{}} 75.40 \\ \footnotesize $\left[\pm 0.40 \right]$\end{tabular} & 
                                                                    \begin{tabular}{@{}c@{}} 60.89 \\ \footnotesize $\left[\pm 0.25 \right]$\end{tabular} & 
                                                                    1.416 & & 
                                                                    \begin{tabular}{@{}c@{}} 58.33 \\ \footnotesize $\left[\pm 0.13 \right]$\end{tabular} & 
                                                                    \begin{tabular}{@{}c@{}} 81.15 \\ \footnotesize $\left[\pm 0.11 \right]$\end{tabular} &  
                                                                    \begin{tabular}{@{}c@{}} 65.54 \\ \footnotesize $\left[\pm 0.31 \right]$\end{tabular} & 
                                                                    1.093 & & 
                                                                    \begin{tabular}{@{}c@{}} 64.98 \\ \footnotesize $\left[\pm 0.47 \right]$\end{tabular} & 
                                                                    \begin{tabular}{@{}c@{}} 84.83 \\ \footnotesize $\left[\pm 0.46 \right]$\end{tabular} & 
                                                                    \begin{tabular}{@{}c@{}} 70.56 \\ \footnotesize $\left[\pm 0.14 \right]$\end{tabular} & 
                                                                    0.896 \\ \midrule
    
$\text{AP}^{'}$                                               & 1 & \begin{tabular}{@{}c@{}} {68.14}  \\ \footnotesize $\left[\pm 0.16 \right]$\end{tabular} & 
                                                                    \begin{tabular}{@{}c@{}} {86.45} \\ \footnotesize $\left[\pm 0.05 \right]$\end{tabular} &  
                                                                    \begin{tabular}{@{}c@{}} {71.18} \\ \footnotesize $\left[\pm 0.10 \right]$\end{tabular} & 
                                                                    \textcolor{blue}{\textbf{1.091}} & & 
                                                                    \begin{tabular}{@{}c@{}} \textbf{62.47} \\ \footnotesize $\left[\pm 1.37 \right]$\end{tabular} & 
                                                                    \begin{tabular}{@{}c@{}} {84.18} \\ \footnotesize $\left[\pm 0.59 \right]$\end{tabular} & 
                                                                    \begin{tabular}{@{}c@{}} \textbf{69.64} \\ \footnotesize $\left[\pm 0.11 \right]$\end{tabular} & 
                                                                    \textbf{1.102} & & 
                                                                    \begin{tabular}{@{}c@{}} 60.91  \\ \footnotesize $\left[\pm 0.47 \right]$\end{tabular} & 
                                                                    \begin{tabular}{@{}c@{}} 82.52 \\ \footnotesize $\left[\pm 0.44 \right]$\end{tabular} &  
                                                                    \begin{tabular}{@{}c@{}} 66.52 \\ \footnotesize $\left[\pm 0.57 \right]$\end{tabular} & 
                                                                    \textbf{1.028} & & 
                                                                    \begin{tabular}{@{}c@{}} 66.95 \\ \footnotesize $\left[\pm 0.04 \right]$\end{tabular} & 
                                                                    \begin{tabular}{@{}c@{}} 85.88 \\ \footnotesize $\left[\pm 0.23 \right]$\end{tabular} & 
                                                                    \begin{tabular}{@{}c@{}} 71.72 \\ \footnotesize $\left[\pm 0.21 \right]$\end{tabular} & 
                                                                    0.916 \\ \midrule

    $\text{MP}^{'}$                                                        & 3 & \begin{tabular}{@{}c@{}} {\textbf{68.76}}  \\ \footnotesize $\left[\pm 0.24 \right]$\end{tabular} & 
                                                                    \begin{tabular}{@{}c@{}} {\textbf{86.47}} \\ \footnotesize $\left[\pm 0.27 \right]$\end{tabular} &  
                                                                    \begin{tabular}{@{}c@{}} {\textbf{71.78}}\\ \footnotesize $\left[\pm 0.29 \right]$\end{tabular} & 
                                                                    \textbf{1.106} & & 
                                                                    
                                                                    \begin{tabular}{@{}c@{}} \textbf{62.47} \\ \footnotesize $\left[\pm 0.13 \right]$\end{tabular} & 
                                                                    \begin{tabular}{@{}c@{}} \textbf{84.66} \\ \footnotesize $\left[\pm 0.84 \right]$\end{tabular} & 
                                                                    \begin{tabular}{@{}c@{}} {69.62} \\ \footnotesize $\left[\pm 0.86 \right]$\end{tabular} & 
                                                                    {1.109} & & 
                                                                    
                                                                    \begin{tabular}{@{}c@{}} {\textbf{60.92}}  \\ \footnotesize $\left[\pm 0.18 \right]$\end{tabular} & 
                                                                    \begin{tabular}{@{}c@{}} \textcolor{blue}{\textbf{82.82}} \\ \footnotesize $\left[\pm 0.11 \right]$\end{tabular} &  
                                                                    \begin{tabular}{@{}c@{}} {\textbf{66.56}} \\ \footnotesize $\left[\pm 0.30 \right]$\end{tabular} & 
                                                                    \textcolor{blue}{\textbf{1.024}} & &
                                                                    
                                                                    \begin{tabular}{@{}c@{}} \textbf{66.66} \\ \footnotesize $\left[\pm 0.24 \right]$\end{tabular} & 
                                                                    \begin{tabular}{@{}c@{}} \textbf{85.77} \\ \footnotesize $\left[\pm 0.03 \right]$\end{tabular} & 
                                                                    \begin{tabular}{@{}c@{}} {\textbf{71.73}} \\ \footnotesize $\left[\pm 0.35 \right]$\end{tabular} & 
                                                                    {\textbf{0.910}} \\ \midrule
    $\text{MP}^{''}$                                                        & 1,5 & \begin{tabular}{@{}c@{}} \textbf{\textcolor{blue}{69.08}}  \\ \footnotesize $\left[\pm 0.23 \right]$\end{tabular} & 
                                                                    \begin{tabular}{@{}c@{}} \textbf{\textcolor{blue}{87.19}} \\ \footnotesize $\left[\pm 0.19 \right]$\end{tabular} &  
                                                                    \begin{tabular}{@{}c@{}} \textbf{\textcolor{blue}{71.98}} \\ \footnotesize $\left[\pm 0.17 \right]$\end{tabular} & 
                                                                    {1.252} & & 
                                                                    
                                                                    \begin{tabular}{@{}c@{}} \textbf{\textcolor{blue}{65.01}} \\ \footnotesize $\left[\pm 0.02 \right]$\end{tabular} & 
                                                                    \begin{tabular}{@{}c@{}} \textbf{\textcolor{blue}{86.60}} \\ \footnotesize $\left[\pm 0.08 \right]$\end{tabular} & 
                                                                    \begin{tabular}{@{}c@{}} \textbf{\textcolor{blue}{71.13}} \\ \footnotesize $\left[\pm 0.21 \right]$\end{tabular} & 
                                                                    \textbf{\textcolor{blue}{1.034}} & & 

                                                                    \begin{tabular}{@{}c@{}} \textbf{\textcolor{blue}{61.67}} \\ \footnotesize $\left[\pm 0.47 \right]$\end{tabular} & 
                                                                    \begin{tabular}{@{}c@{}} \textbf{82.75} \\ \footnotesize $\left[\pm 0.17 \right]$\end{tabular} & 
                                                                    \begin{tabular}{@{}c@{}} \textbf{\textcolor{blue}{67.38}} \\ \footnotesize $\left[\pm 0.47 \right]$\end{tabular} & 
                                                                    1.091 & & 
                                                                    
                                                                    \begin{tabular}{@{}c@{}} \textbf{\textcolor{blue}{67.63}} \\ \footnotesize $\left[\pm 0.16 \right]$\end{tabular} & 
                                                                    \begin{tabular}{@{}c@{}} \textbf{\textcolor{blue}{86.20}} \\ \footnotesize $\left[\pm 0.06 \right]$\end{tabular} & 
                                                                    \begin{tabular}{@{}c@{}} \textbf{\textcolor{blue}{72.61}} \\ \footnotesize $\left[\pm 0.01 \right]$\end{tabular} & 
                                                                    \textbf{\textcolor{blue}{0.902}}  \\
    
                                                                    \bottomrule                                                     
\end{tabular}}
\caption{Image retrieval performance of AdvProp-D (AP$^{'}$) and MDProp (MP) methods against standard training (ST) and adversarial training (AT) when used with \textit{ResNet18}, \textit{ResNet50} in \textit{S2SD}\cite{roth2021simultaneous}, and \textit{ResNet152} architectures on the \textit{CUB200}\cite{cub200} dataset. Acronyms and adversarial data generation parameters are the same as in Table \ref{table:1}.}
\label{table:r18_r152_r50s2sd_results}
\end{center}
\end{table*}
\setlength{\tabcolsep}{1.4pt}

\paragraph{Effect of Separate BN Layers.}
From the results in Table \ref{table:1} and Table \ref{table:r18_r152_r50s2sd_results}, we confirm that the use of adversarial training, which does not use separate BN layers, can result in enhanced adversarial robustness compared to the standard training, but it always reduces the clean data performance. We also experimented with training a DML model with mixed inputs of clean and MTAX data while not using separate BN layers\footref{footnote:all_tables} and found that mixed inputs without separate BNs do not provide clean data performance gains, and robustness stays similar to adversarial training. Therefore, confirming AdvProp's \cite{advprop} hypothesis of input distribution shift handling by separate BN layers.

\paragraph{Input Distribution Shift by MTAXs.}
The gradient descent directions during MTAX generation are constrained towards overlapped embedding spaces of the model under training. Whereas the optimization complexity of the STAX generation process is relatively low because of a bigger feasible solution space. This leads us to hypothesize that since the generation of MTAXs follows a different process, they are differently distributed than STAXs, as well as the clean data.

To demonstrate that MTAXs are differently distributed, we follow the methodology of Xie et al. \cite{advprop}. We compared the trained model's performance when additional BN layers were used during inference instead of the main BN layers. While evaluating the trained ResNet50 model with Multisimilarity loss in the MDProp framework using three BN layers for clean, STAX, and MTAX inputs, respectively, 
We found that the additional BN layers for STAXs and MTAXs result in a mean $0.7\%$ and $1.8\%$ decrease in R@1 scores for the test CUB200 data.

\begin{figure}[!h]
\centering
\begin{subfigure}[t]{\linewidth}
    \centering
    \includegraphics[width=0.40\linewidth]{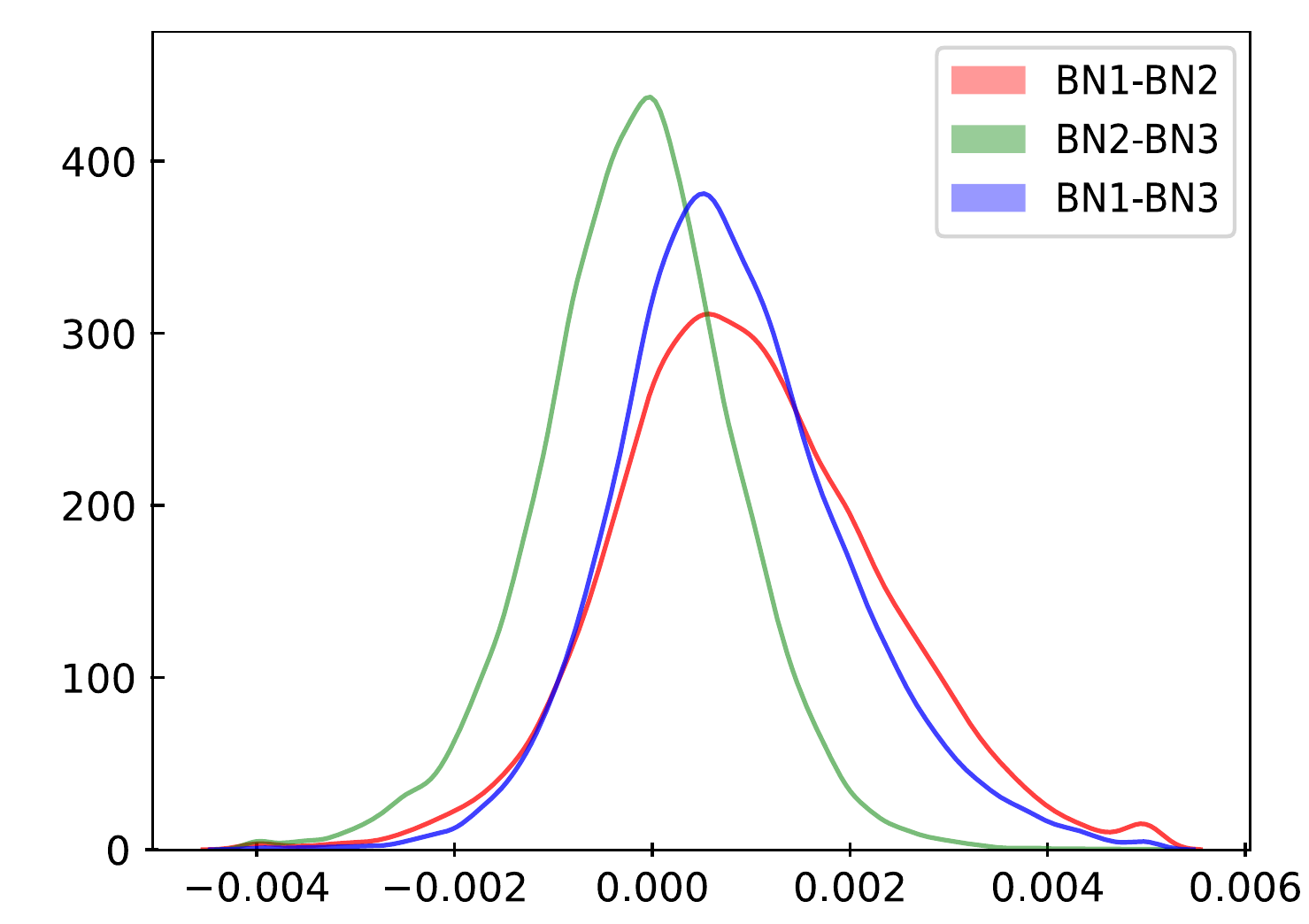}
    \qquad
    \includegraphics[width=0.40\linewidth]{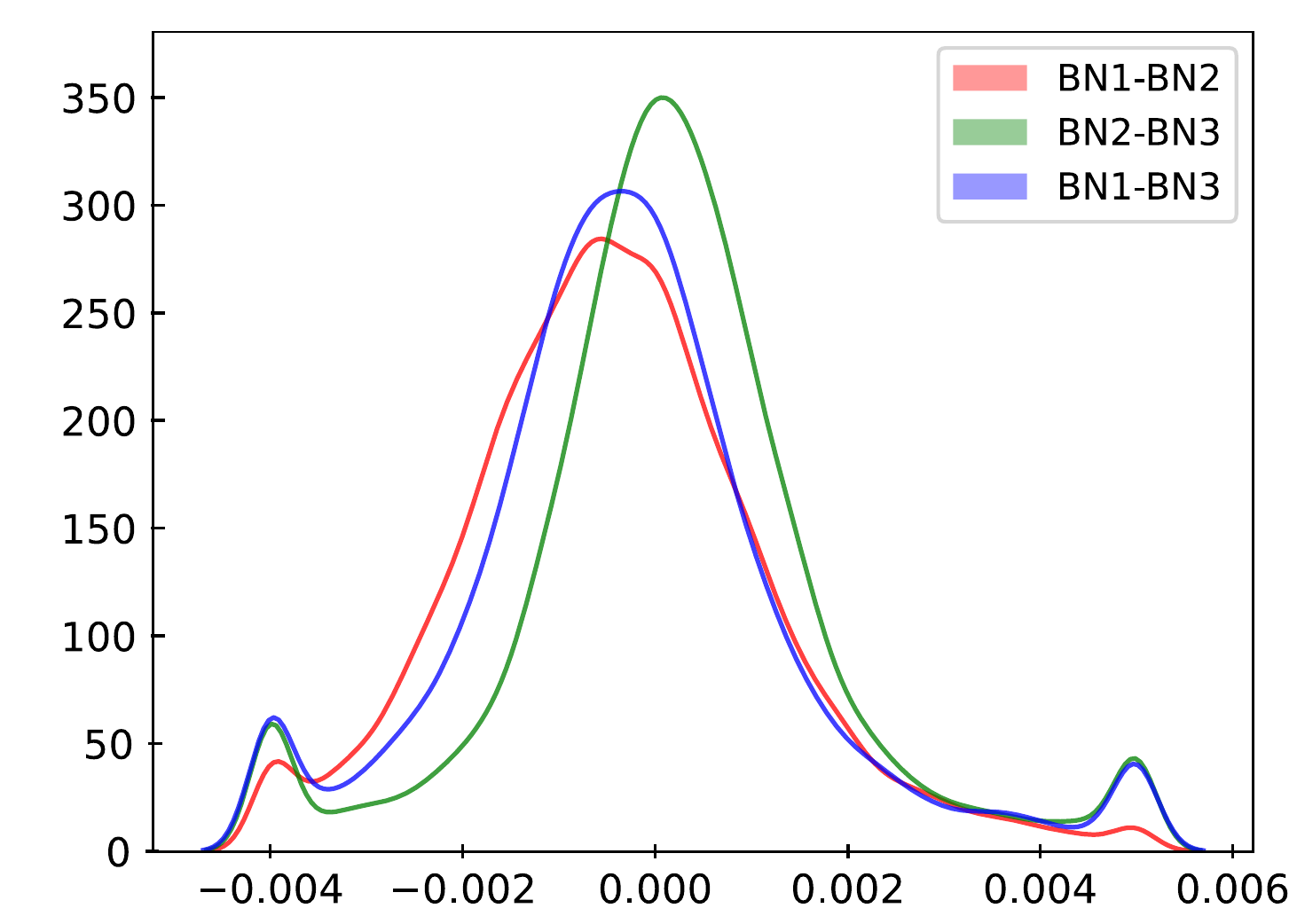}
\end{subfigure}
\caption{Layer-wise variations in the learned parameters of the BN layers used for different types of inputs in MDProp. BN1, BN2, and BN3 represent the BN layers used for the clean data, STAXs, and MTAXs, respectively. Every difference was recorded at the same depth for each BN layer pair.
A shift from zero with a significant variation indicates considerably different learned parameters in the pair of BN layers.
}
\label{fig:dist_plots}
\end{figure}

We also compared the pairwise difference in the learned $\beta$ and $\gamma$ parameters of the BN layers used for different kinds of training data. As illustrated in Fig. \ref{fig:dist_plots}, we found a clear significant pairwise variation between these parameters of the different BN layers used for clean, STAX, and MTAX inputs. This difference in the learned parameters further proves that these BN layers were trained for different input distributions meaning MTAXs follow different distributions than clean data and STAXs, thereby causing input distribution shift and requiring additional BN layers during training. 

\setlength{\tabcolsep}{2pt}
\begin{table}[t]
\begin{center}
\resizebox{\columnwidth}{!}{
\begin{tabular}{lcrrrrrrrrrr}
    \toprule
    \multirow{2}{*}{Method}& \multirow{2}{*}{$T$}& \multicolumn{4}{c}{Clean Data} & & \multicolumn{4}{c}{Adversarial Data}\\\cmidrule{3-6} \cmidrule{8-11}
     &  & R@1 & R@4 & NMI & $\pi_{ratio}$ & & R@1 & R@4 & NMI & $\pi_{ratio}$\\ \toprule
    \rowcolor{gray!10}ST                           & - &  {78.09}  & {86.55} & {89.98} & {0.469} && 54.60 & 66.09 & 84.96 & 0.636  \\ \midrule
    
    AP$^{'}$                                        & 1 & 77.36  & 86.17 & {89.98} & \textcolor{blue}{\textbf{0.410}} &&  \textbf{71.96} & {82.86} & {88.57} & \textcolor{blue}{\textbf{0.422}}  \\ \midrule
    
    $\text{MP}^{'}$                                                       & 3 & \textbf{77.73}  & {\textbf{86.98}} & {\textbf{89.99}} & 0.475 &&  \textcolor{blue}{\textbf{72.95}} & \textcolor{blue}{\textbf{83.90}} & \textcolor{blue}{\textbf{88.82}} & {0.456}  \\ \midrule

    $\text{MP}^{''}$                                                       & 1,5 & \textbf{\textcolor{blue}{78.70}}  & \textcolor{blue}{\textbf{87.19}} & \textcolor{blue}{\textbf{90.27}} & \textbf{0.452} &&  {71.89} & {\textbf{83.00}} & {\textbf{88.62}} & \textbf{0.438}  \\ \bottomrule
   
\end{tabular}}
\caption{Results for the SOP \cite{sop} dataset while using \textit{ResNet50} with \textit{Multisimilarity} \cite{multisimilarity} loss. Adversarial data was generated following Table \ref{table:1}.}
\label{table:sop_results}
\end{center}
\end{table}
\setlength{\tabcolsep}{1.4pt}

\section{Related Work}
\noindent \textbf{DML}. Conventional DML techniques are based on improved ranking losses, data sampling methods, data augmentation, and some extensions to the DML. Our work can be seen as an extension of standard DML techniques that augment different types of adversarial data in a disentangled learning environment to improve DML performance against multiple input distributions. 

\noindent \textbf{Training with AXs}. The clean data performance degradation of adversarial training \cite{kurakin2018adversarial,pgd,xie2019feature,free_adv_training_shafahi2019,fast_adv_training_andriushchenko2020} is well-known mainly for the DL models of moderate sizes. Tsipras et al. \cite{tsipras2018there} argue that the performance tradeoff between accuracy and robustness is inevitable, attributing this phenomenon to learning fundamentally different feature representations. 

Xie et al. \cite{advprop} proposed using separate BN layers for clean and adversarial data during training to improve clean data accuracy of the classification DL models and achieved significant gains in the clean data accuracy for the ImageNet \cite{imagenet} data. To further revamp the AdvProp's costly training in the classification setting, Mei et al. \cite{fastadvprop} propose the \textit{FastAdvProp} method that changes fractions of the clean and adversarial training data and then uses the free adversarial training technique \cite{free_adv_training_shafahi2019}. In the semi-supervised learning domain, Raghunathan et al. \cite{rst} proposed the robust self-training (RST) method to improve the clean accuracy and robustness. However, there exists no work focusing on improving the image retrieval performance of DML models on clean and adversarial data simultaneously.

This paper focuses on improving the image retrieval performance of DML models on clean and adversarial data simultaneously. We show that further increasing the number of separate BN layers with different types of input data in our MDProp can further improve the performance as long as the additional data has meaningful features. We also provide the first extension of AdvProp \cite{advprop} in the DML setting. We also demonstrate a practical methodology to utilize existing pre-trained parameters of conventional architectures to enable transfer learning for the auxiliary BN layer parameters, thus optimizing computational complexity.

\section{Conclusion}
In this paper, we proposed \textit{MDProp} to improve: (1) DML models' image retrieval performance for clean input and (2) robustness against multiple distributions different than clean input, specifically adversarial attacks. \textit{MDProp} generated MTAXs along with STAXs while leveraging disentangled learning during training to regularize \textit{overlapped} embedding space of DML models, thereby resulting in enhanced generalization. \textit{MDProp} can be used with a manifold of architectures, loss functions, distillation-based approaches, and datasets to further increase image retrieval performance on inputs following multiple distributions. In particular, MDProp increased clean data Recall@1 scores by $2.95\%$ and multi-distribution input robustness by $2.12$ times for the ResNet-based state-of-the-art models. The \textit{AdvProp-D} case of \textit{MDProp} provides a DML extension of the proven \textit{AdvProp} method \cite{advprop}.

{\small
\bibliographystyle{ieee_fullname}
\bibliography{egbib}
}

\newpage
\appendix

\section{Evaluation Metrics}
We use the standard evaluation metrics in deep metric learning (DML): Recall@K (R@K) \cite{recall} with $k=\{1,4\}$, Normalized Mutual Information (NMI) \cite{nmi}, and $\pi_{ratio}$. Increased \textit{R@k} and \textit{NMI} values indicate improved image retrieval performance and clustering quality, respectively, and the decreased $\pi_{ratio}$ values approximately indicate increased inter-class and decreased intra-class distances in the embedding space of the trained model.

\subsection{Recall@k \cite{recall}}
For a given DML function $f$, let $\mathcal{F}_{q}^{k}$ be the set of first $k$ nearest neighbors of a sample $x_q \in \mathcal{X}_{test}$ defined as
\begin{equation}
\mathcal{F}_{q}^{k}=\underset{\mathcal{F} \subset \mathcal{X}_{\text {test }},|\mathcal{F}|=k}{\arg \min } \sum_{x_{n} \in \mathcal{F}} d\left(f\left(x_{q}\right), f\left(x_{n}\right)\right)
\end{equation}
Finally, Recall@k is calculated as
\begin{equation}
R @ k=\frac{1}{\left|\mathcal{X}_{\text {test }}\right|} \sum_{x_{q} \in \mathcal{X}_{\text {test }}} \begin{cases}1 & \exists x_{i} \in \mathcal{F}_{q}^{k} \text { s.t. } y_{i}=y_{q} \\ 0 & \text { otherwise }\end{cases}
\end{equation}
This means Recall@k measures the average number of cases in which, for a given query $x_q$, there is at least one sample among its top $k$ nearest neighbors $x_i$ with the same class, i.e., $y_i=y_q$.

\subsection{Normalized Mutual Information (NMI) \cite{nmi}}
NMI quantifies the clustering quality in the embedding space of a DML model $f$. To calculate NMI for the embedding space $\Phi_{X_{test}}$ of all test samples $x_i \in \mathcal{X}_{test}$, we assign a cluster label $w_i$ corresponding to each sample $x_i$ indicating the closest cluster center and define $\Omega = \{\omega_k\}_{k=1}^{K}$ with $\omega_k = \{i|w_i = k\}$ and $K = |\mathcal{C}|$ being the number of classes and clusters. Similarly for the true labels $y_i$ we define $\Upsilon = \{\upsilon_c\}^{K}_{c=1}$ with $\upsilon_c = \{i|y_i = c\}$. The NMI is then computed with mutual Information $I(\cdot,\cdot)$ between cluster and labels, and entropy $H(\cdot,\cdot)$ on the clusters and labels, respectively, as
\begin{equation}
NMI(\Omega, \Upsilon)=\frac{I(\Omega, \Upsilon)}{2(H(\Omega)+H(\Upsilon))}
\end{equation}

\subsection{Embedding Space Density ($\pi_{ratio}$)}
We define embedding space density $\pi_{ratio}$ as 

\begin{equation}
    \pi_{ratio}(\Phi) = \frac{\pi_{intra}(\Phi)}{\pi_{inter}(\Phi)}
\end{equation}
 where $\pi_{intra}(\Phi)$ is the intra class distance and $\pi_{inter}(\Phi)$ inter class distance in the feature space $\Phi_{\mathcal{X}}:=\{f_{\theta}(x) \mid x \in \mathcal{X}_{test}\}$ of a DML model $f_{\theta}$ and they are calculated as follows:

\begin{equation}
\pi_{intra}(\Phi) = \frac{1}{Z_{\text {intra }}} \sum_{y_{l} \in \mathcal{Y}} \sum_{\phi_{i}, \phi_{j} \in \Phi_{y_{l}}, i \neq j} d\left(\phi_{i}, \phi_{j}\right)
\end{equation}

 \begin{equation}
\pi_{inter}(\Phi) = \frac{1}{Z_{\text {inter }}} \sum_{y_{l}, y_{k}, l \neq k} d\left(\mu\left(\Phi_{y_{l}}\right), \mu\left(\Phi_{y_{k}}\right)\right)
\end{equation}

\noindent Here, $\Phi_{y_{l}}=\left\{\phi_{i}:=f_{\theta}\left(x_{i}\right) \mid x_{i} \in \mathbb{X}, y_i=y_l \right\}$ denotes the set of embedded samples of a class $y_l$. $\mu(\Phi_{y_l})$ their mean embedding and $Z_{intra}$, $Z_{inter}$ are the normalization constants.

\section{Benchmarks}
We evaluate the performance on the CUB200 \cite{cub200}, CARS 196 \cite{cars}, and Stanford Online Products \cite{sop} benchmarks following the experimental setting by Roth et al. \cite{roth2021simultaneous} for data pre-processing.

\noindent\textbf{CUB200} \cite{cub200} contains 200 bird classes over 11,788 images, whereas the first and last 100 classes with 5864/5924 images are used for training and testing, respectively.

\noindent\textbf{CARS196} \cite{cars} contains 196 car classes and 16,185 images, where again, the first and last 98 classes with 8054/8131 images are used to create the training/testing split.

\noindent\textbf{Stanford Online Products (SOP)} \cite{sop} is built around 22,634 product classes over 120,053 images and contains a provided split: 11318 selected classes with 59551 images are used for training, and 11316 classes with 60502 images for testing.

\section{Complete Experimental Setup}
For reproducibility, we present the complete experimental details used to evaluate the performance of the MDProp. For all experiments, we followed the setup used by Roth et al. \cite{roth2021simultaneous} except for the frozen batch normalization \cite{bn}. We use frozen batch normalization only for the baselines to reproduce the results and make a fair comparison with state-of-the-art methods. Our setup includes the ResNet18, ResNet50, and ResNet152 architectures \cite{resnet} with normalization of the output embeddings with dimensionality 128 and optimization with Adam \cite{adam} using a learning rate of $10-5$ and weight decay of $4 \cdot 10-4$. The input images were randomly resized and cropped from the original image size to $224 \times 224$ pixels for training. Further augmentation by random horizontal flipping with $p = 0.5$ is applied. During the testing, center crops of size $224 \times 224$ were used. The batch size was set as 112. Training runs on CUB200 and CARS 196 were performed over 150 epochs and 100 epochs for the SOP for all experiments without any learning rate schedule.

We also used S2SD \cite{roth2021simultaneous} with ResNet50 architecture and Multisimilarity loss \cite{multisimilarity}. We retained the remaining hyperparameters in the S2SD default, as shown in Table 1 of the paper by Roth et al. \cite{roth2020revisiting}. Everything was implemented in PyTorch \cite{pytorch}. The experiments were performed on GPU servers containing \textit{Nvidia Tesla V100}, \textit{Titan V}, and \textit{RTX 1080Ti}s. However, double memory usage is required in our methods. For cases exceeding the memory requirement, then a single GPU's available VRAM, we used data parallelization to distribute the training on multiple GPUs to meet memory requirements. Each result in Table 1 in the paper is averaged over three seeds; for Table 2, two seeds are used. We report commonly neglected means and standard deviations for reproducibility and validity.

For adversarial example (AX) generation during training, we used the projected gradient descent (PGD) update \cite{pgd} to generate single and multi-targeted adversarial examples (MTAXs). We set the number of iterations in PGD to 1, $L_{\infty}$ constraint $\epsilon$ on the adversarial noise to $0.01$, and the PGD learning rate as $\epsilon/\text{Attack Iterations}$. We take four different values $T=2,3,5,10$ as attack targets during the MTAX generation. The loss function was kept squared $L_2$ norm for generating feature space AXs. For the robustness assessment of the MDProp's trained DML models against inputs following multiple distributions during inference, we generated single as well as multi-targeted AXs. 

\subsection{DML Loss Functions}
\subsubsection{Multisimilarity \cite{multisimilarity}}
Multisimilarity loss \cite{multisimilarity} uses the concept of different types of similarities in all positive and negative samples for an anchor $x_i$ in training data while using hard sample mining:
\begin{equation}
d_{c}^{*}(i, j)= \begin{cases}d_{c}\left(\psi_{i}, \psi_{j}\right) & d_{c}\left(\psi_{i}, \psi_{j}\right)>\min _{j \in \mathcal{P}_{i}} d_{c}\left(\psi_{i}, \psi_{j}\right)-\epsilon \\ d_{c}\left(\psi_{i}, \psi_{j}\right) & d_{c}\left(\psi_{i}, \psi_{j}\right)<\max _{k \in \mathcal{N}_{i}} d_{c}\left(\psi_{i}, \psi_{k}\right)+\epsilon \\ 0 & \text { otherwise }\end{cases}
\end{equation}
\begin{equation}
\begin{aligned}
\mathcal{L}_{m}=\frac{1}{b} \sum_{i \in \mathcal{B}}\left[\frac{1}{\alpha} \log \left[1+\sum_{j \in \mathcal{P}_{i}} \exp \left(-\alpha\left(d_{c}^{*}\left(\psi_{i}, \psi_{j}\right)-\lambda\right)\right)\right]\right] \\
+\sum_{i \in \mathcal{B}}\left[\frac{1}{\beta} \log \left[1+\sum_{k \in \mathcal{N}_{i}} \exp \left(\beta\left(d_{c}^{*}\left(\psi_{i}, \psi_{k}\right)-\lambda\right)\right)\right]\right]
\end{aligned}
\end{equation}
Where $d_c$ is the cosine similarity, and $\mathcal{P}_i/\mathcal{N}_i$ is the set of positives and negatives for $x_i$ in the mini-batch, respectively. We use the default values $\alpha = 2$, $\beta = 40$, $\lambda = 0.5$ and $\epsilon = 0.1$.

\subsubsection{ArcFace \cite{deng2019arcface}}
Arcface transforms the standard softmax formulation typically used in classification problems to retrieval-based problems by enforcing an angular margin between the embeddings $\phi = f(\mathcal{X})$ and an approximate center $W \in \mathbb{R}^{c\times d}$ for each class.
We used additive angular margin penalty $\gamma = 0.5$. The radius of the effectively utilized hypersphere $\mathbb{S}$ denoted as the scaling $s = 16$ was used. The class centers were optimized with a learning rate of 0.0005.

\subsection{Adversarial Training with Targeted Attacks}
It is well known that adversarial training results in highly robust models, but causes a reduction in the clean data performance of the model. In this study, our primary focus is to improve the accuracy of clean data using AXs in the form of multi-distribution inputs. Hence, to make the comparison fair and effectively evaluate the effect of separate BN layers, we used both clean and adversarial data during training without using separate BN layers. For generating adversarial data, we use the same single targeted AXs $x_{adv}^{t}$ used in the AdvProp-D case of MDProp, which are generated as
\begin{equation}\label{eq:feature_space_ax}
    x_{adv}^f = x_{i}^{j} + \delta_{f}^{t} \ \ s.t. \ \ \delta_{f}^{t} = \argmin_{||\delta||_{\infty}\leq\epsilon}\left[\mathcal{L}(f(x_{i}^{j}+\delta),f(x_{i}^{k}))\right]
\end{equation}
\noindent where $\mathcal{L}$ measures the distance, $f$ is the DML model, and $x_{i}^{k}$ is the target identity's image. 

Finally, the objective of the adversarial training in our setting is as follows:
\begin{equation}\label{eq:advprop_d_objective}
    \mathcal{Z}_1 = \argmin_{\theta}  \left[\mathbb{E}_{\left\{\substack{(x,y)\sim \mathbb{D} \\ \delta_{f}^{t} \sim \mathbb{D}^{'}}\right\}} \mathcal{L}\left(\theta, \left(x,y\right), \left(x+\delta_{f}^{t},y\right)\right)\right]
\end{equation}
\noindent where $(x,y)\sim \mathbb{D}$ denotes a clean data instance. $\mathcal{L}$ denotes the DML training loss. $\theta=\{\theta_{n},\theta_{b}\}$ are the parameters of the model that does not have auxiliary BN layers.

\subsection{Evaluating Multi-Distribution Inputs}
For robustness assessment, the STAX and MTAX datasets were generated corresponding to the clean samples in the test sets of the CUB200 \cite{cub200}, CARS 196 \cite{cars}, and SOP \cite{sop} datasets. We used the \textit{PGD} \cite{pgd} update with 20 iterations, calling it \textit{PGD-20} attacks. We used 0.01 and 0.1 for the $\epsilon$ constraint. for MTAXs, we used $T=5$. The remaining attack hyper-parameters were kept the same as during the training time of attack generation.

\section{Detailed Results}
This section presents the detailed results of the comparison of our methods against baselines on clean data performance in Table \ref{table:clean_performance}, robustness against STAX inputs in Table \ref{table:stax_robustness}, robustness against powerful STAX inputs generated using $\epsilon=0.1$ in Table \ref{table:strong_stax_robustness}, clean data performance and adversarial robustness across architectures and SOTA S2SD methods in Table \ref{table:multi_arch_s2sd}, and clean data performance for larger models with \textit{\textbf{larger embedding dimensions}} in Table \ref{table:r152_512}. Each table also presents the results for the case where MTAXs \textbf{\textit{without separate batch normalization}} were used, which is included in the adversarial training method case. In addition, these tables show the results for \textit{\textbf{additional values of the number of targets}} $T$ for MTAX generation.

\subsection{Performance on MTAX Inputs}
We also evaluate the performance against MTAX inputs to test check decreased overlapped feature space in the MDProp models. The results for MTAX inputs are presented in Table \ref{table:mtax_robustness}. Clearly, MDProp models result in improved metrics for MTAX inputs. 

\subsection{Effect of Number of Adversarial Targets $T$}
\begin{figure}[t]
    \centering
    \includegraphics[width=\columnwidth]{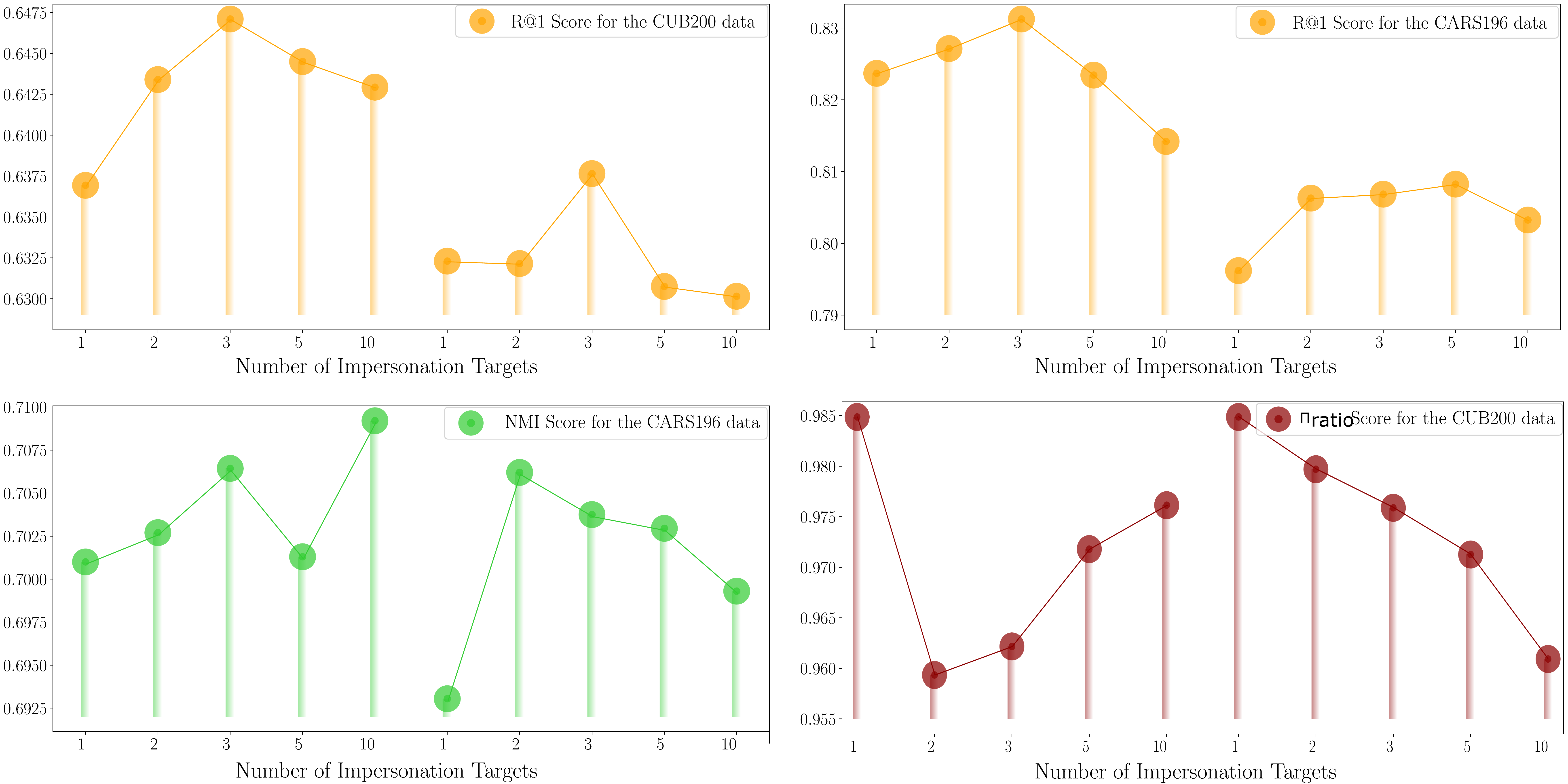}
    \caption{Impact of the number of attack targets $T$ on the clean data performance. The sample trends generally demonstrate \textit{improved} R@1, NMI, and $\pi_{ratio}$ scores with the increase in $T$ initially and then the decrease due to increased MTAX generation complexity and restricted attack generation procedure.}
    \label{fig:tar_num_impact}
\end{figure}

Figure \ref{fig:tar_num_impact} illustrates the effect of $T$ parameters on the performance of the trained model using MDProp. We conducted experiments using five values of $T$: 1, 2, 3, 5, and 10. It was found that increasing $T$ improves performance on clean data only up to a certain number for which the predefined \textit{generation} recipe's hyperparameters provide sufficient semantic capability to the attack generation procedure, causing the positions in the embedding space of generated MTAXs shift to the overlapped regions of the DML model under training. In particular, MDProp using clean and MTAXs performed best for $T=3$, and MDProp using clean, STAXs, and MTAXs performed best for $T=5$.
For smaller values of $T$, lesser performance improvements result because of the decreased probability of finding highly overlapped embedding-space regions.

\subsection{Results for PGD-20 attacks with $\epsilon=0.1$}
To evaluate the robustness gains for powerful attacks, we generate attacks with larger values of the $\epsilon$ constraint. We use $\epsilon=0.1$ for generating single targeted AXs to compare the reduction in performance of AdvProp-D and MDProp. Similar to the case for PGD-20 attacks with $\epsilon=0.01$, robustness gain was found to be marginally higher for the AdvProp-D followed by MDProp, which can be seen in Table \ref{table:strong_stax_robustness}. AdvProp-D and MDProp result in significantly high adversarial robustness compared to the baseline standard training and the adversarial training methods. Hence, we can conclude that our proposed AdvProp-D and the MDProp methods provide significant robustness gains for attacks of varying strength with different sizes of adversarial noise.

\subsection{Results When MDProp Use 4 Separate BN Layers}
Table \ref{table:4bn_clean_performance} presents the results when MDProp uses three additional BN layers for the STAXs and MTAXs data generated for two different numbers of targets. Clearly, there were significant performance gains. However, the performance gains remained marginally lower than those of MDProp using the three separate BN layers presented in the paper.

\begin{figure}[t]
    \centering
    \begin{subfigure}{0.48\columnwidth}
        \centering
        \includegraphics[width=\linewidth]{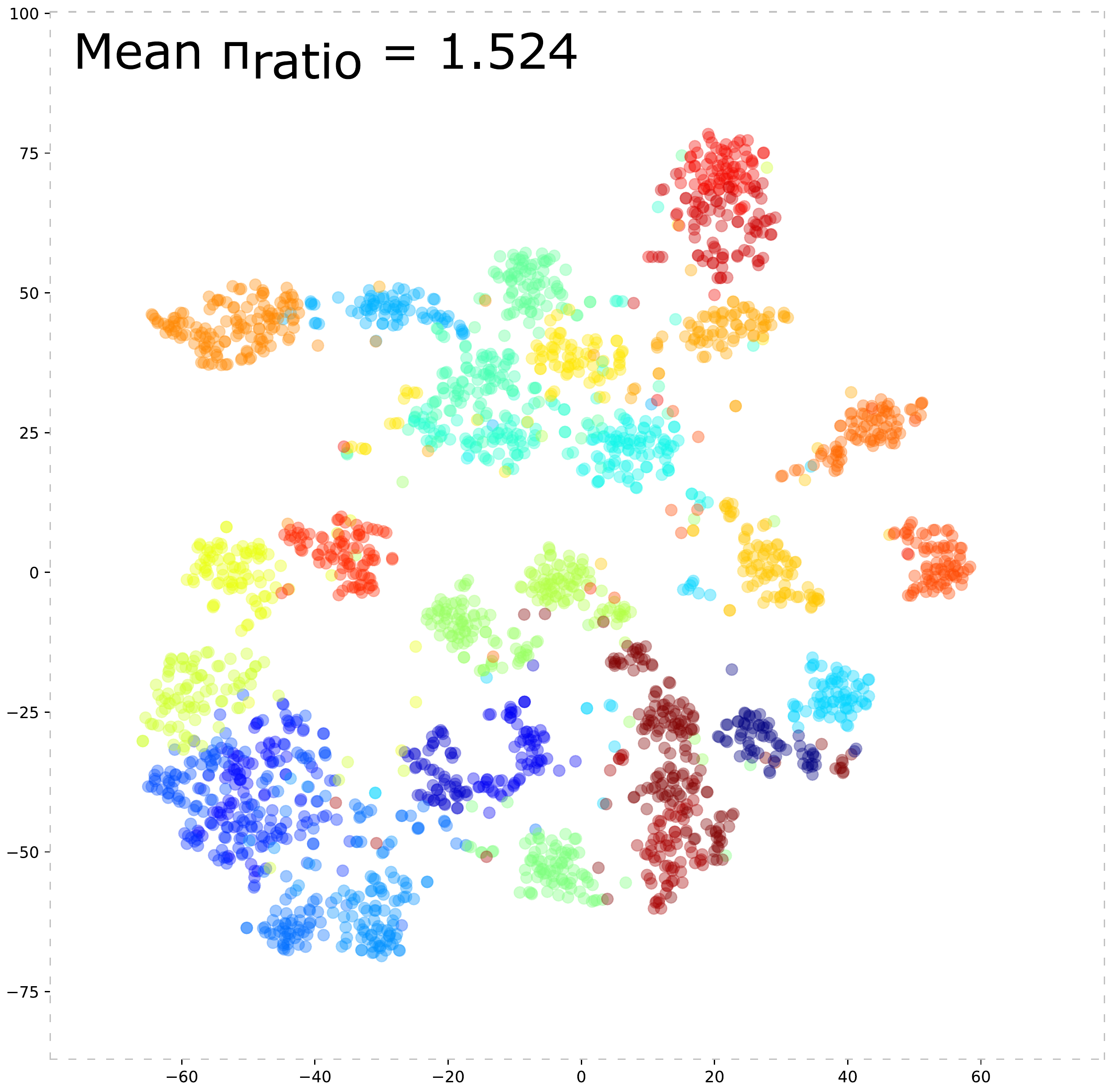}
        \subcaption{Standard Training}
    \end{subfigure}
    \hfill
    \begin{subfigure}{0.48\columnwidth}  
        \centering 
        \includegraphics[width=\linewidth]{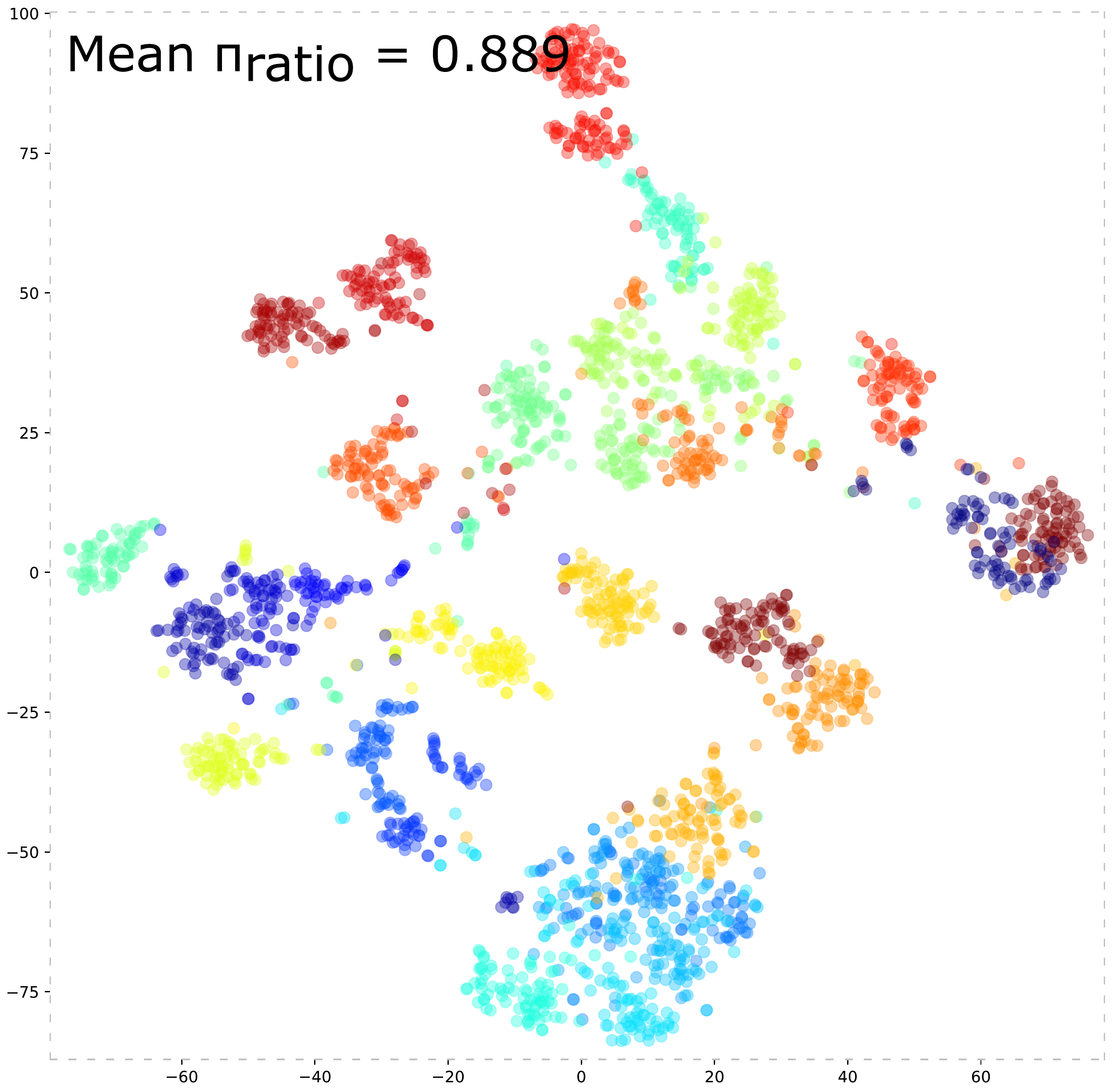}  
        \subcaption{MasterProp}
    \end{subfigure}
    \caption{t-SNE \cite{tsne} visualization of embedding space of DML models trained using (a) \textit{standard training} and (b) \textit{MDProp} on the \textit{CARS196} dataset \cite{cars}. The decreased mean $\pi_{ratio}$ score for MasterProp means sparser embedding space.}
    \label{fig:inter and intra class separation}
\end{figure}

\setlength{\tabcolsep}{2pt}
\begin{table*}[!ht]
\begin{center}
\resizebox{\textwidth}{!}{
 & 
                                                                    0.699  \\ \bottomrule
\end{tabular}}
\caption{Detailed results for the clean data performance of our AdvProp-D (AP$^{'}$) MDProp (MP) methods against baseline standard training (ST) \cite{roth2021simultaneous} and adversarial training (AT). $MP^{'}$ and $MP^{''}$ represent one and two additional BN layers, respectively. The performance was evaluated for the models trained using multisimilarity \cite{multisimilarity} and ArcFace \cite{deng2019arcface} losses. Compared to Table 1 in the paper, this table additionally demonstrates the results for the models trained using multiple MTAX targets $T$ and the effect of using the separate batch normalization layers along with separate BN layers by showing the additional results for the use of MTAXs in the AT setting, i.e., without using separate batch normalization layers ($T=2,3,5,10$ for method AT in the table).}
\label{table:clean_performance}
\end{center}
\end{table*}
\setlength{\tabcolsep}{1.4pt}

\setlength{\tabcolsep}{2pt}
\begin{table*}[!ht]
\begin{center}
\resizebox{\textwidth}{!}{
 & 
                                                                    0.686  \\ \bottomrule
\end{tabular}}
\caption{Detailed results for the adversarial data performance of our AdvProp-D (AP$^{'}$) and our MDProp (MP) methods against baseline standard training (ST) \cite{roth2021simultaneous} and adversarial training (AT)  when single targeted PGD-20 attacks were generated using $\epsilon=0.01$. The performance was evaluated for the models trained using multisimilarity \cite{multisimilarity} and ArcFace \cite{deng2019arcface} losses. Compared to Table 1 in the paper, this table additionally demonstrates the results for the robustness of the models trained using multiple MTAX targets$T$ and the effect of using the separate batch normalization layers along with separate BN layers by showing the additional results for the use of MTAXs in the AT setting, i.e., without using separate batch normalization layers ($T=2,3,5,10$ for method AT in the table).}
\label{table:stax_robustness}
\end{center}
\end{table*}
\setlength{\tabcolsep}{1.4pt}

\setlength{\tabcolsep}{2pt}
\begin{table*}[!ht]
\begin{center}
\resizebox{\textwidth}{!}{
 & 
                                                                    1.004  \\ \bottomrule
\end{tabular}}
\caption{Detailed results for the adversarial data performance of our AdvProp-D (AP$^{'}$) and MDProp (MP) methods against the baseline standard training (ST) and adversarial training (ST) \cite{roth2021simultaneous} when the \textbf{\textit{ stronger}} single-targeted PGD-20 attacks were generated using $\epsilon=0.1$. The performance is evaluated for the models trained using the Multisimilarity \cite{multisimilarity} and ArcFace \cite{deng2019arcface} losses. Compared to Table 1 in the paper, this table additionally demonstrates the results for the robustness of the models trained using multiple MTAX targets $T$ and the effect of using separate batch normalization layers along with separate BN layers by showing the additional results for the use of MTAXs in the AT setting, that is, without using separate batch normalization layers ($T=2,3,5,10$ for method AT in the table).}
\label{table:strong_stax_robustness}
\end{center}
\end{table*}
\setlength{\tabcolsep}{1.4pt}

\setlength{\tabcolsep}{2pt}
\begin{table}[!ht]
\begin{center}
\resizebox{0.7\linewidth}{!}{
\begin{tabular}{lcrrrr}
    \toprule
    Method& $T$& R@1 & R@4 & NMI & $\pi_{ratio}$ \\ \toprule
    
    ST                           & - &  \begin{tabular}{@{}c@{}} 36.35  \\ \footnotesize $\left[\pm 0.41 \right]$\end{tabular} & 
                                                                    \begin{tabular}{@{}c@{}} 62.23 \\ \footnotesize $\left[\pm 0.87 \right]$\end{tabular} &  
                                                                    \begin{tabular}{@{}c@{}} 47.69 \\ \footnotesize $\left[\pm 0.42 \right]$\end{tabular} & 
                                                                    1.447 \\ \midrule
    
    \multirow{10}{*}{AT} & 1 &  \begin{tabular}{@{}c@{}} 55.31  \\ \footnotesize $\left[\pm 0.70 \right]$\end{tabular} & 
                                                                    \begin{tabular}{@{}c@{}} 82.40 \\ \footnotesize $\left[\pm 0.25 \right]$\end{tabular} &  
                                                                    \begin{tabular}{@{}c@{}} 67.83 \\ \footnotesize $\left[\pm 0.22 \right]$\end{tabular} & 
                                                                    0.755 \\ \cmidrule{2-6}
    ~                                                        & 2 & \begin{tabular}{@{}c@{}} 53.76  \\ \footnotesize $\left[\pm 0.92 \right]$\end{tabular} & 
                                                                    \begin{tabular}{@{}c@{}} 81.62 \\ \footnotesize $\left[\pm 0.29 \right]$\end{tabular} &  
                                                                    \begin{tabular}{@{}c@{}} 66.84 \\ \footnotesize $\left[\pm 0.26 \right]$\end{tabular} & 
                                                                    0.757 \\ \cmidrule{2-6}
    ~                                                        & 3 & \begin{tabular}{@{}c@{}} 54.16  \\ \footnotesize $\left[\pm 1.01 \right]$\end{tabular} & 
                                                                    \begin{tabular}{@{}c@{}} 82.43 \\ \footnotesize $\left[\pm 0.19 \right]$\end{tabular} &  
                                                                    \begin{tabular}{@{}c@{}} 67.21 \\ \footnotesize $\left[\pm 0.66 \right]$\end{tabular} & 
                                                                    0.775  \\ \cmidrule{2-6}
    ~                                                        & 5 & \begin{tabular}{@{}c@{}} 54.72  \\ \footnotesize $\left[\pm 0.17 \right]$\end{tabular} & 
                                                                    \begin{tabular}{@{}c@{}} 81.91 \\ \footnotesize $\left[\pm 0.64 \right]$\end{tabular} &  
                                                                    \begin{tabular}{@{}c@{}} 67.73 \\ \footnotesize $\left[\pm 0.54 \right]$\end{tabular} & 
                                                                    0.758  \\ \cmidrule{2-6}
    ~                                                        & 10 & \begin{tabular}{@{}c@{}} 54.32  \\ \footnotesize $\left[\pm 0.22 \right]$\end{tabular} & 
                                                                    \begin{tabular}{@{}c@{}} 81.66 \\ \footnotesize $\left[\pm 0.25 \right]$\end{tabular} &  
                                                                    \begin{tabular}{@{}c@{}} 67.74 \\ \footnotesize $\left[\pm 0.15 \right]$\end{tabular} & 
                                                                    0.765  \\ \midrule
    
    AP$^{'}$                                        & 1 & \begin{tabular}{@{}c@{}} 59.97  \\ \footnotesize $\left[\pm 0.17 \right]$\end{tabular} & 
                                                                    \begin{tabular}{@{}c@{}} 83.83 \\ \footnotesize $\left[\pm 1.02 \right]$\end{tabular} &  
                                                                    \begin{tabular}{@{}c@{}} 71.27 \\ \footnotesize $\left[\pm 1.74 \right]$\end{tabular} & 
                                                                    0.746  \\ \midrule
    
    \multirow{8}{*}{$\text{MP}^{'}$}                                      & 2 & \begin{tabular}{@{}c@{}} 61.00  \\ \footnotesize $\left[\pm 0.53 \right]$\end{tabular} & 
                                                                    \begin{tabular}{@{}c@{}} 86.43 \\ \footnotesize $\left[\pm 0.50 \right]$\end{tabular} &  
                                                                    \begin{tabular}{@{}c@{}} 72.14 \\ \footnotesize $\left[\pm 0.76 \right]$\end{tabular} & 
                                                                    0.612 \\ \cmidrule{2-6}
    ~                                                        & 3 & \begin{tabular}{@{}c@{}} 61.13  \\ \footnotesize $\left[\pm 0.85 \right]$\end{tabular} & 
                                                                    \begin{tabular}{@{}c@{}} 86.05 \\ \footnotesize $\left[\pm 0.41 \right]$\end{tabular} &  
                                                                    \begin{tabular}{@{}c@{}} 71.88 \\ \footnotesize $\left[\pm 0.61 \right]$\end{tabular} & 
                                                                    0.617 \\ \cmidrule{2-6}
    ~                                                        & 5 &  \begin{tabular}{@{}c@{}} 60.65  \\ \footnotesize $\left[\pm 0.54 \right]$\end{tabular} & 
                                                                    \begin{tabular}{@{}c@{}} 86.19 \\ \footnotesize $\left[\pm 0.55 \right]$\end{tabular} &  
                                                                    \begin{tabular}{@{}c@{}} 71.92 \\ \footnotesize $\left[\pm 0.58 \right]$\end{tabular} & 
                                                                    0.632 \\ \cmidrule{2-6}
    ~                                                        & 10 & \begin{tabular}{@{}c@{}} 60.55  \\ \footnotesize $\left[\pm 2.49 \right]$\end{tabular} & 
                                                                    \begin{tabular}{@{}c@{}} 85.76 \\ \footnotesize $\left[\pm 1.39 \right]$\end{tabular} &  
                                                                    \begin{tabular}{@{}c@{}} 71.45 \\ \footnotesize $\left[\pm 0.29 \right]$\end{tabular} & 
                                                                     0.618 \\ \midrule
    \multirow{8}{*}{$\text{MP}^{''}$}                                      & 1,2 & \begin{tabular}{@{}c@{}} 62.69  \\ \footnotesize $\left[\pm 0.00 \right]$\end{tabular} & 
                                                                    \begin{tabular}{@{}c@{}} 86.96 \\ \footnotesize $\left[\pm 0.45 \right]$\end{tabular} &  
                                                                    \begin{tabular}{@{}c@{}} 72.68 \\ \footnotesize $\left[\pm 0.81 \right]$\end{tabular} & 
                                                                    0.621 \\ \cmidrule{2-6}
    ~                                                        & 1,3 & \begin{tabular}{@{}c@{}} 62.04  \\ \footnotesize $\left[\pm 0.07 \right]$\end{tabular} & 
                                                                    \begin{tabular}{@{}c@{}} 86.34 \\ \footnotesize $\left[\pm 0.18 \right]$\end{tabular} &  
                                                                    \begin{tabular}{@{}c@{}} 72.81 \\ \footnotesize $\left[\pm 0.10 \right]$\end{tabular} & 
                                                                    0.606 \\ \cmidrule{2-6}
    ~                                                        & 1,5 &  \begin{tabular}{@{}c@{}} 61.41  \\ \footnotesize $\left[\pm 0.38 \right]$\end{tabular} & 
                                                                    \begin{tabular}{@{}c@{}} 86.49 \\ \footnotesize $\left[\pm 0.17 \right]$\end{tabular} &  
                                                                    \begin{tabular}{@{}c@{}} 72.33 \\ \footnotesize $\left[\pm 0.75 \right]$\end{tabular} & 
                                                                    0.624 \\ \cmidrule{2-6}
    ~                                                        & 1,10 & \begin{tabular}{@{}c@{}} 60.11  \\ \footnotesize $\left[\pm 1.56 \right]$\end{tabular} & 
                                                                    \begin{tabular}{@{}c@{}} 86.38 \\ \footnotesize $\left[\pm 0.28 \right]$\end{tabular} &  
                                                                    \begin{tabular}{@{}c@{}} 71.37 \\ \footnotesize $\left[\pm 0.33 \right]$\end{tabular} & 
                                                                     0.619  \\ \bottomrule
\end{tabular}}
\caption{Detailed results for the adversarial data performance of our AdvProp-D (AP$^{'}$) and our MDProp (MP) methods against the baseline standard training (ST) \cite{roth2021simultaneous} and adversarial training (AT)  when \textbf{white-box multi-targeted PGD-20 attacks with $T=5$} were generated using $\epsilon=0.01$. The performance was evaluated for models trained using multisimilarity \cite{multisimilarity}  loss on CUB200 \cite{cub200} data.}
\label{table:mtax_robustness}
\end{center}
\end{table}
\setlength{\tabcolsep}{1.4pt}

\setlength{\tabcolsep}{2pt}
\begin{table*}[!ht]
\begin{center}
\resizebox{\textwidth}{!}{
}
\caption{Detailed results demonstrating clean data performance and robustness gains by the AdvProp-D and MDProp methods across architectures (ResNet18, ResNet50, and ResNet152) of different sizes on the CUB200 \cite{cub200} dataset. ResNet50 was implemented using the SOTA S2SD method \cite{roth2021simultaneous}.}
\label{table:multi_arch_s2sd}
\end{center}
\end{table*}
\setlength{\tabcolsep}{1.4pt}

\setlength{\tabcolsep}{2pt}
\begin{table*}[ht!]
\begin{center}
\resizebox{0.65\textwidth}{!}{
%
 & 
                                                                    1.078\\  \bottomrule
\end{tabular}}
\caption{The results of our MDProp (MP) method when used with three additional BN layers on the CUB200 \cite{cub200} and CARS 196 \cite{cars} datasets using multisimilarity \cite{multisimilarity} loss. $T$ represents the number of attack targets used for different types of adversarial data generation. The performance is evaluated for the models trained using the Multisimilarity \cite{multisimilarity}  loss on the CUB200 \cite{cub200} data.}
\label{table:4bn_clean_performance}
\end{center}
\end{table*}
\setlength{\tabcolsep}{1.4pt}

\setlength{\tabcolsep}{7pt}
\begin{table}[!ht]
\begin{center}
\resizebox{\columnwidth}{!}{
\begin{tabular}{lcrrrrrrr}
    \toprule
     Method & $T$ & R@1 & R@4 & NMI & $\pi_{ratio}$\\ \toprule
    ST                           & - &  64.97 & 85.15 & 66.52 & 1.163\\ \midrule
    
    \multirow{5}{*}{AT} & 1 &  65.20 &  84.65 & 66.59 & 1.348 \\ \cmidrule{2-6}
    ~                                                        & 2 &  64.87 & 84.49 &  66.89 & 1.352 \\ \cmidrule{2-6} 
    ~                                                        & 3 &  65.03 & 84.78 & 66.76 & 1.352 \\ \cmidrule{2-6}
    ~                                                        & 5 &  65.23 & 84.88 & 67.09 & 1.354 \\ \cmidrule{2-6}
    ~                                                        & 10 & 65.30 & 84.92 & 66.81 & 1.346 \\ \midrule
    
    AP$^{'}$                                        & 1 & 68.85 & 68.91 & 68.33 & 1.214 \\ \midrule
    
    \multirow{5}{*}{MP}                                      & 2 & 68.39 & 86.56 & 69.30 & 1.211 \\ \cmidrule{2-6}
    ~                                                        & 3 & 68.91 & 86.69 & 68.85 & 1.181 \\ \cmidrule{2-6}
    ~                                                        & 5 &  68.19 & 86.36 & 68.63 & 1.190 \\ \cmidrule{2-6} 
    ~                                                        & 10 & 68.62 & 86.95 & 69.24 & 1.170 \\ \bottomrule
   
\end{tabular}}
\caption{Results when an embedding size of 512 was used while training the ResNet152 architecture with the S2SD \cite{roth2021simultaneous} method on the CUB200 \cite{cub200} dataset. MDProp, followed by AdvProp-D, demonstrated significant clean data performance gains over the baselines.}
\label{table:r152_512}
\end{center}
\end{table}
\setlength{\tabcolsep}{1.4pt}

\end{document}